\documentclass{article}

\usepackage{microtype}
\usepackage{graphicx}
\usepackage{subcaption}
\usepackage{booktabs} % for professional tables

\usepackage{hyperref}

\usepackage[preprint]{icml2026}

\usepackage{amsmath,amsfonts,bm}

\def\eqref#1{equation~\ref{#1}}
\def\1{\bm{1}}

\def\vs{{\bm{s}}}

\DeclareMathAlphabet{\mathsfit}{\encodingdefault}{\sfdefault}{m}{sl}
\SetMathAlphabet{\mathsfit}{bold}{\encodingdefault}{\sfdefault}{bx}{n}

\usepackage{amsmath}
\usepackage{amssymb}
\usepackage{mathtools}
\usepackage{amsthm}
\usepackage{enumitem}
\usepackage{pifont}

\usepackage[capitalize,noabbrev]{cleveref}

\theoremstyle{plain}

\theoremstyle{definition}

\theoremstyle{remark}

\usepackage{xspace}
\DeclareRobustCommand\onedot{\ifx\@let@token.\else.\null\fi\xspace}
\def\eg{\emph{e.g}\onedot} 
\def\ie{\emph{i.e}\onedot} 
 
\def\etc{\emph{etc}\onedot} \def\vs{\emph{vs}\onedot}
\def\wrt{w.r.t\onedot}

\makeatother

\definecolor{mygreen}{RGB}{34,139,34}
\definecolor{mylightblue}{RGB}{0,162,230}
\definecolor{mylightred}{RGB}{208,86,99}
\definecolor{deepyellow}{RGB}{255, 215, 0} 
\definecolor{mygray}{RGB}{232, 232, 232}

\newcommand{\model}{LINA\xspace}

\makeatletter
\def\blfootnote#1{\xdef\@thefnmark{}\@footnotetext{\scriptsize #1}}
\makeatother

\usepackage[textsize=tiny]{todonotes}

\icmltitlerunning{LINA: Linear Autoregressive Image Generative Models with Continuous Tokens}

\begin{document}

\twocolumn[
  \icmltitle{LINA: Linear Autoregressive Image Generative Models with Continuous Tokens}



  \icmlsetsymbol{equal}{*}

  \begin{icmlauthorlist}
  \icmlauthor{Jiahao Wang}{hku}
  \icmlauthor{Ting Pan}{ucas}
  \icmlauthor{Haoge Deng}{ucas}
  \icmlauthor{Dongchen Han}{thu}
  \icmlauthor{Taiqiang Wu}{hku}
  \icmlauthor{Xinlong Wang}{baai}
  \icmlauthor{Ping Luo}{hku}
  \end{icmlauthorlist}

  \icmlaffiliation{hku}{The University of Hong Kong}
  \icmlaffiliation{ucas}{University of Chinese Academy of Sciences}
  \icmlaffiliation{thu}{Tsinghua University}
  \icmlaffiliation{baai}{Beijing Academy of Artificial Intelligence}

  \icmlcorrespondingauthor{Xinlong Wang}{wangxinlong@baai.ac.cn}
  \icmlcorrespondingauthor{Ping Luo}{pluo@cs.hku.hk}

  \icmlkeywords{Machine Learning, ICML}

  \vskip 0.3in
]



\printAffiliationsAndNotice{}  % no special notice (required even if empty)

\begin{abstract}

Autoregressive models with continuous tokens represent a unique paradigm for visual generation, showing strong promise in text-to-image (T2I) synthesis but suffering from heavy computational costs. In this work, we investigate how compute-efficient linear attention should be designed within this framework. 
We start with a systematic empirical study to examine how linear attention scales with parameter counts under different design choices.
Specifically, we examine two key design choices: (i) \textit{normalization paradigms} in linear attention—division-based \vs subtraction-based, and (ii) the use of a \textit{depthwise convolution} on image features for locality modeling augmentation. 
The scaling results indicate that while subtraction-based normalization is effective for image classification, division-based normalization is more amenable to linear generative transformers; besides, convolutions play a key role in linear attention for autoregressive modeling, consistent with prior findings in diffusion models.
Furthermore, we explore introducing \textit{gating mechanisms}, a key design choice in causal linear attention, into bidirectional linear attention, and as a result, we propose a \textit{KV gate}. 
By applying data-independent learnable parameters to the key and value states, our method assigns token-wise weights to memory, enabling flexible memory management, similar to the forget gate in language models.
% 
% We validate its impact on generation quality via detailed ablation studies using FID, and further visualize the patterns it learns.
% 
Building on these designs, we offer \model, a simple and compute-efficient text-to-image generative model with pure linear attention, capable of rapidly generating high fidelity 1024$\times$1024 images from user instructions. 
\model achieves strong results on both class-conditional and T2I generation. Compared to diffusion models of similar scale and autoregressive models with softmax attention, \model delivers competitive performance—FID of 2.18 on ImageNet ($\sim$1.4B) and an overall score of 0.74 on GenEval ($\sim$1.5B). For efficiency, a single linear attention module reducing FLOPs by $\sim$61\% over softmax attention. 
Code and models are released at \url{https://github.com/techmonsterwang/LINA}.

\end{abstract}
\section{Introduction}
\label{sec:introduction}

\begin{figure*}[htbp]
    \centering
    \includegraphics[width=0.89\textwidth]{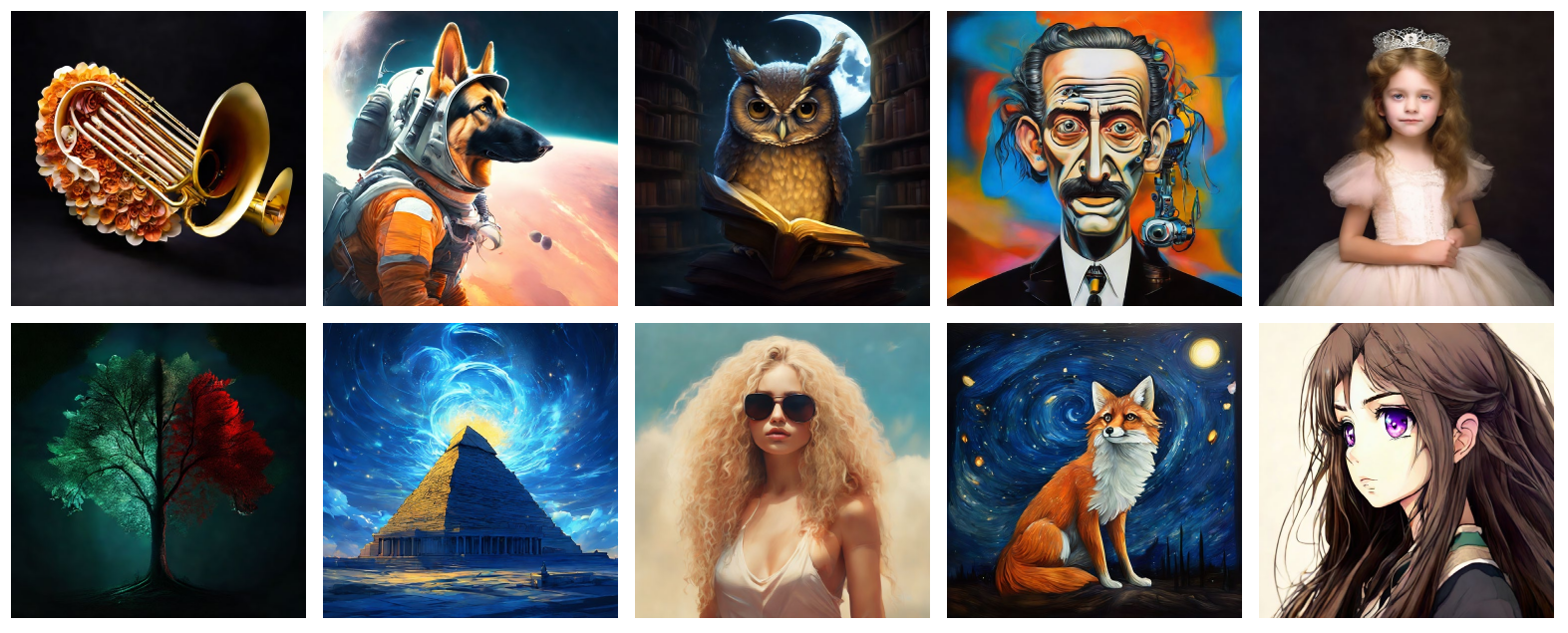}
    % \vspace{-3em}
    \caption{Qualitative results of 1024px samples powered by \model.
    }
    \vspace{-1em}
    \label{fig:teaser}
\end{figure*}

The field of image generation is evolving rapidly~\cite{ho2020denoising, peebles2023scalable, chen2024pixart1, chen2024pixart3}. 
Autoregressive models with continuous tokens~\cite{li2024autoregressive} emerge as a competitive alternative to diffusion models and show strong promise, having been validated in text-to-image tasks~\cite{fan2024fluid}, inspiring video generation~\cite{deng2024autoregressive}, and scaling up to large models (\eg, 14B parameters~\cite{team2025nextstep}).
The paradigm involves both multi-step autoregression and diffusion, and a major bottleneck lies in its efficiency: its reliance on quadratic computation complexity softmax attention~\cite{vaswani2017attention} makes it less practical for long-sequence generation such as high-resolution images or long videos.

Linear attention~\cite{katharopoulos2020transformers}, as a natural alternative, has been extensively explored in both vision transformer (ViT)~\cite{dosovitskiy2020image}-based perception models~\cite{cai2023efficientvit, han2023flatten} and diffusion transformer (DiT)~\cite{peebles2023scalable}-based generative models~\cite{xie2024sana, xie2025sana, wang2025lit}.
However, it remains unclear how linear attention should be adapted to autoregressive generative models.
Unlike DiTs, which generate tokens in parallel, autoregressive models perform inference in a sequential manner: image tokens are generated step by step, with the set of known tokens gradually expanding as the inference progresses.
Such distinction suggest that the design choices for linear attention may need a careful reconsideration, \eg, normalization paradigms~\cite{han2024bridging}, and gating mechanisms~\cite{yang2023gated, yang2024gated} commonly used in autoregressive language models.

In this paper, we systematically study what suitable design choices of linear attention fit autoregressive image generative models.
Our approach builds on the NOVA~\cite{deng2024autoregressive} framework as the baseline, and we analyze linear-attention design choices for ImageNet~\cite{deng2009imagenet} $256 \times 256$ class-conditional image (C2I) generation.
To start with, we conduct an empirical study on the \textit{scaling behavior \wrt parameter counts}, focusing on two main factors: \textit{normalization paradigm} and \textit{locality augmentation} of linear attention (Sec.~\ref{sec:scaling}). 
The underlying reason is straightforward. Normalization enforces the attention weights to sum to one, which stabilizes the scale of activations and influences training dynamics. Meanwhile, linear attention, compared with softmax attention, is well known to suffer from insufficient locality modeling~\cite{han2023flatten, han2024bridging}. 
Thus, we introduce two design choices: division-based versus subtraction-based normalization for linear attention, and whether image tokens should be augmented with locality. For each setting, we train models at three model capacities: $\sim$0.4B, $\sim$0.6B, and $\sim$1.4B parameters.
% 
% We use FID~\cite{heusel2017gans}, sFID~\cite{nash2021generating}, Inception Score~\cite{salimans2016improved}, and Precision/Recall~\cite{kynkaanniemi2019improved} as evaluation metrics, and report how sampling quality scales with parameter counts. 
% 
Counterintuitively, the injectivity brought by subtraction-based normalization turns out to be \textit{not} essential for autoregressive image generative modeling, despite its emphasized role in classification models~\cite{han2024bridging}. Meanwhile, the inductive bias~\cite{cordonnier2019relationship} introduced by depthwise convolution (DWC) for locality enhancement does \textit{not} appear to be restrictive when scaling up parameters.

% Next, we explore the gating mechanism~\cite{yang2023gated, yang2024gated, zhang2024gated, lin2025forgetting, qiu2025gated}

Then, we explore the gating mechanism~\cite{yang2023gated, yang2024gated, zhang2024gated, lin2025forgetting, qiu2025gated}—commonly used in autoregressive language models—for image generative modeling (Sec.~\ref{sec:gating}).
In prior work, gating has been shown to flexibly manage the memory in \textit{causal} linear attention. However, a key challenge is how to adapt it to the \textit{bidirectional} attention used in image generation, where we develop a simple yet effective \textit{KV gate} method.
KV gate are \textit{data-independent}, scalar-valued \textit{learnable parameters} with \textit{no} explicit range constraints. 
By applying independent gating factors to the key and value, they enable flexible memory management and fits naturally with bidirectional attention. 
We conduct detailed ablation studies on their different modes and find that proper management of \textit{both the memory and the normalization term} is crucial. 
We further provide detailed visualizations of the KV gate to see what it has learned. 
% As a result, our KV gate can serve as a readily applicable practice for linear image generative transformers. 

\begin{figure*}[!t]
    \centering
    \includegraphics[width=0.92\linewidth]{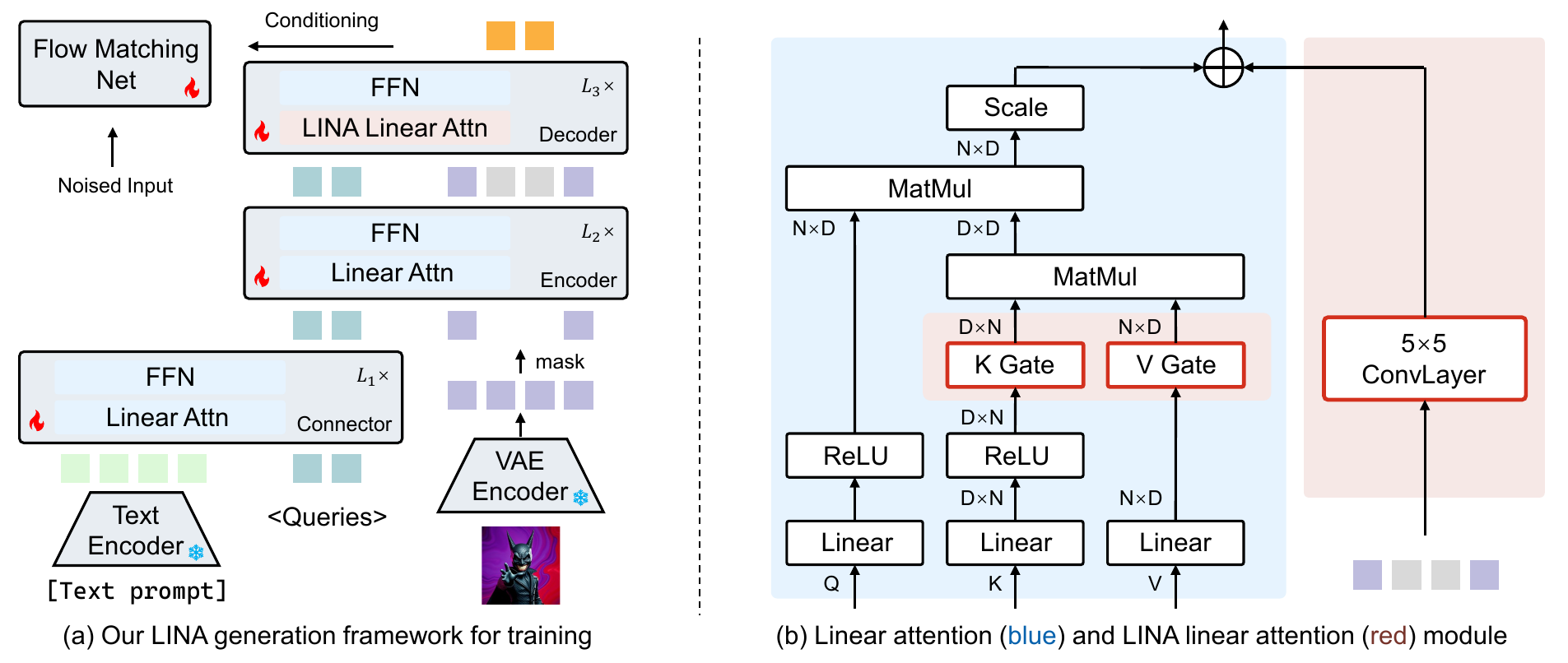}
    \caption{\textbf{Overview of \model}: Fig.~(a) illustrates the training pipeline, with a \textit{Connector} for extracting text information, an \textit{Encoder} to extract unmasked tokens, and a \textit{Decoder} to reconstruct masked tokens for conditioning. A denoising flow matching MLP is used to sample tokens. Fig.~(b) shows the \textit{division-based normalization} linear attention, and our introduced DWC module and KV gate (Sec.~\ref{sec:method}).   
    }
    \vspace{-1em}
    \label{fig:model_training}
\end{figure*}

Lastly, based on these explorations, we propose \model, a compute-efficient linear autoregressive image generative model (Sec.~\ref{sec:experiments}). 
\model employs \textit{division-based normalization} linear attention enhanced with a \textit{DWC module} for locality enhancement, together with the \textit{KV gate}.
We first validate \model on class-conditional image generation, achieving an FID of 2.18 on the ImageNet 256$\times$256 benchmark, which is competitive with SOTA diffusion models.
We further extend \model to text-to-image generation, where it follows user instructions and efficiently produces high-fidelity images up to 1024px, achieving a highly competitive GenEval score of 74.
In terms of efficiency, linear attention in our \model reduces FLOPs by $\sim$61\% compared with softmax attention when generating 1024px images, demonstrating the computational advantage of our approach.

Through LINA, we seek to unlock the potential of linear attention and call on the community to develop efficient transformer alternatives for broad applications.

\section{Related Work}
\label{sec:related_work}

Here we briefly review related work. A detailed version is provided in Appendix.~\ref{sec:appendix.related}. Our study builds on autoregressive models with continuous tokens~\citep{li2024autoregressive, fan2024fluid}, which have become a mainstream approach in image generation in recent years. This line of research has been explored not only with 14B-parameter autoregressive models (\eg, NextStep-1~\citep{team2025nextstep}), but also successfully extended to video generation (\eg, NOVA~\citep{deng2024autoregressive}).
One of the key bottlenecks of such methods lies in their computational efficiency. To this end, we draw inspiration from efficient linear attention~\citep{katharopoulos2020transformers, choromanski2020rethinking}, which has already been successfully applied to both LLMs~\citep{yang2023gated} and visual perception task~\citep{cai2023efficientvit, han2023flatten}. Along this line, several studies~\citep{xie2024sana, xie2025sana, wang2025lit, zhu2024dig, pu2024efficient} have studied designing eficient DiTs with linear attention. Differently, we thoroughly discuss how linear attention should be designed in autoregressive image generative models. 
\section{Preliminary}
\label{sec:preliminary}

\subsection{Autoregressive Modeling with Continuous Tokens} 
\label{sec:preliminary.mar}

Given a target of $N$ tokens $\{X_1, \dots, X_{N}\}$ to predict, masked autoregressive models (MAR)~\citep{li2024autoregressive} complete the prediction task in $K$ steps.
Illustrated in Fig. 2-(a), at every step $k$, a random-order autoregressive model predicts a set of tokens 
$\mathbf{S}_k = \{X_i, X_{i+1}, \dots, X_j\}$ with $\cup_k{\mathbf{S}_k}=\{X_1, \dots, X_{N}\}$, conditioned on the previously generated tokens $\{X_1, \dots, X_{i-1}\}$: 

\begin{equation}
\small
\begin{aligned}
p(X_1, \dots, X_{N}) &= p(\mathbf{S}_1, \dots, \mathbf{S}_{K}) \\
&= \prod^{K}_{k}
p( \mathbf{S}_k ~ |~ \mathbf{S}_1, \dots, \mathbf{S}_{k-1}).
\label{eq:masked_autoregressive_model}
\end{aligned}
\end{equation}

Typically, MAR models consist of a network (\eg, Transformer~\citep{vaswani2017attention}) that predicts a condition vector from the input, and a diffusion~\citep{ho2020denoising, sohl2015deep} or flow matching ~\citep{lipman2022flow, liu2022flow} head (\eg, MLP) that models the next token distribution conditioned on this vector.
% The head employs either diffusion~\citep{ho2020denoising, sohl2015deep} or flow matching methods~\citep{lipman2022flow, liu2022flow}, enabling the model to operate on continuous tokens.

Efficiency is a practical challenge for MAR framework, as both the autoregressive and diffusion processes require multiple iterations. 
In this work, we take a closer look at the design choices of linear attention in this context, focusing on scaling behavior with parameter and gating mechanisms.

\subsection{Linear Attention} 
\label{sec:preliminary.la}
Given an input sequence $I\in\mathbb{R}^{N\times D}$ of length $N$, we denote the queries, keys, and values in linear attention~\citep{katharopoulos2020transformers} by $Q, K, V\in\mathbb{R}^{N\times D}$. 
We refer to the kernel function as $\phi(\cdot)$ (\eg, $\texttt{ReLU}$) and the output as $O\in\mathbb{R}^{N\times D}$ (for simplicity, we assume attention head as 1).  
Linear attention introduces a normalization factor $\gamma\in\mathbb{R}^{N}$ that ensures the normalized attention weights for each token $i$ sum to $1$. Based on how this normalization term is defined, we categorize linear attention as follows\footnote{Here our terminology is based on~\citep{han2024bridging} and~\citep{fan2025rectifying}.}.

\paragraph{Division-based normalization.} In this formulation, the normalization factor in linear attention is placed in the denominator, akin to the \texttt{softmax} operation in full attention~\citep{vaswani2017attention}:

\begin{equation}
\small
\begin{aligned}
O^{(\texttt{d})}_i 
= \sum_{j=1}^{N} \frac{A^{(\texttt{d})}_{ij}}{{\gamma^{(\texttt{d})}_i}} V_j
&= \sum_{j=1}^{N}\frac{\phi(Q_i)\phi(K_j)^\top}{\sum_{m=1}^{N}{\phi(Q_i) \phi(K_m)^\top}}V_j\\          
&= \frac{\phi(Q_i) \left(\sum_{j=1}^{N}{\phi(K_j)^\top V_j }\right)}{\phi(Q_i) \left(\sum_{m=1}^{N}{\phi(K_m)^\top}\right)},
\label{eq:division_based_linear}
\end{aligned}
\end{equation} 

where ${\gamma^{(\texttt{d})}_i}\in\mathbb{R}$ denotes the scalar-valued division-based normalization term, calculated from $Q_i\in\mathbb{R}^{1\times D}$ and the set of key states $K_m\in\mathbb{R}^{1\times D}$ for $m \in [1, N]$. 
Linear attention reduces the computational complexity from $\mathcal{O}(N^2)$ in full attention to $\mathcal{O}(N)$, 
since for every query $Q_i$, both the memory $M=\sum_{j=1}^{N}{\phi(K_j)^\top V_j }\in\mathbb{R}^{D\times D}$ and $z=\sum_{m=1}^{N}{\phi(K_m)^\top} \in\mathbb{R}^{D\times 1}$ are shared and thus need to be computed only once.

\paragraph{Subtraction-based normalization.} In this form, the normalization term is introduced as a distinct term, imparting linear attention with an injective property~\citep{han2024bridging}:  

% \begin{equation} 
% \begin{aligned}
% \small
% O^{(\texttt{s})}_i
% &= \sum_{j=1}^{N} \left(A^{(\texttt{\texttt{s}})}_{ij}-{\gamma^{(\texttt{s})}_i}\right)V_j         
% = \sum_{j=1}^{N}\left[\phi(Q_i)\frac{1}{N}\phi(K_j)^\top -{\left(     \frac{1}{N}\sum_{m=1}^{N} \phi(Q_i)\frac{1}{N}\phi(K_m)^\top - \frac{1}{N}\right) }\right]V_j \\         
% &= \phi(Q_i) \left(\frac{1}{N}\sum_{j=1}^{N}{\phi(K_j)^\top V_j  }\right) -{ \left(\phi(Q_i)  \frac{1}{N}\sum_{m=1}^{N}{\phi(K_m)^\top} - 1\right)\frac{1}{N} } \sum_{j=1}^{N}V_j,  
% \label{eq:subtraction_based_linear}
% \end{aligned}
% \end{equation} 

\begin{equation}
\small
\begin{aligned}
O^{(\texttt{s})}_i
&= \sum_{j=1}^{N} \left(A^{(\texttt{\texttt{s}})}_{ij}-{\gamma^{(\texttt{s})}_i}\right)V_j         
\\         
&= \phi(Q_i) \left(\frac{1}{N}\sum_{j=1}^{N}{\phi(K_j)^\top V_j  }\right)\\
&-{ \left(\phi(Q_i)  \frac{1}{N}\sum_{m=1}^{N}{\phi(K_m)^\top} - 1\right)\frac{1}{N} } \sum_{j=1}^{N}V_j,  
\label{eq:subtraction_based_linear}
\end{aligned}
\end{equation} 

where $\gamma^{(\texttt{s})}_i\in\mathbb{R}$ denotes the scalar-valued subtraction-based normalization term. Injective property allows linear attention (Eq.~\ref{eq:subtraction_based_linear}) to distinguish different queries, which is similar to softmax attention. 
% In comparison, in contrast to the non-injective nature of Eq.~\ref{eq:division_based_linear}. 

\section{Scaling Behavior of Linear Attention}
\label{sec:scaling}

% \subsection{\model Motivation and Roadmap} 
% \label{sec:method.roadmap}

% \paragraph{Motivation.}

% 
Despite linear attention has shown promise in DiTs~\citep{pu2024efficient, xie2024sana, xie2025sana} and ViTs~\citep{cai2023efficientvit, han2023flatten, guo2024slab}, its suitable design for autoregressive image generation task remains unclear in the literature, inspiring us to find suitable design choices of linear attention to save computation while retaining generative performance.
% In this work, we seek to find suitable design choices of linear attention to save computation while retaining generative performance.
% 
In this section, we focus on the scaling behavior with respect to parameter counts of linear attention and highlight two core design choices.
\textbf{Q1} (\textit{linear attention paradigm choice}): Which paradigm—division-based or subtraction-based normalization—better supports parameter scaling?
\textbf{Q2} (\textit{locality choice}): Without softmax, linear attention shows limited ability to model local patterns, which may affect performance. Does introducing a locality inductive bias affect parameter scaling?
% Beyond scaling, we further draw inspiration from gating mechanisms in autoregressive language models to study memory control in our context.
% \textbf{Q3} (\textit{memory management}): Linear attention has a fixed memory size to process image tokens without clear differentiation. As tokens vary in semantics, long sequences may cause memory collisions~\citep{yang2024gated, schlag2021linear}, hindering the focus on informative tokens.

To answer this question, we build our model upon the NOVA~\citep{deng2024autoregressive} framework, but rigorously replaces all softmax attention with linear attention. As shown in Fig.~\ref{fig:model_training} (the inference framework are provided in Appendix~\ref{sec:appendix.inference}), 
the model consists of:
a \textit{Connector} for integrating text or class information;
an \textit{Encoder} and \textit{Decoder} for predicting conditioning; and
a denoising flow matching network (\ie, a small MLP)~\citep{li2024autoregressive} for modeling token probability distributions. 
% Then, we explore the design of linear attention blocks. 
We conduct a systematic scaling study to compare different linear attention paradigms and the effect of introducing a DWC module for locality modeling enhancement.

% \begin{figure}[t]
% \centering
% \hspace{-0.5em}
% \begin{minipage}{0.49\linewidth}{
% \includegraphics[width=0.99\textwidth]{figure/dwc_mar_compare.pdf}
% }\end{minipage}
% % \hfill
% \begin{minipage}{0.5\linewidth}{
% \caption{
% \textbf{DWC helps locality.} 
% (a) A random-order autoregressive model with bidirectional attention predicts next tokens based on the predicted tokens. 
% When the target token (\eg, the 8th) is surrounded by predicted tokens (\eg, the 3rd), the model faces challenges due to the limited local modeling capacity. 
% (b) By using convolution layer, DWC module gathers information from nearby known tokens when predicting the current token, thereby facilitating linear attention. 
% }
% \label{fig:dwc_linear}
% }\end{minipage}
% \vspace{-1em}
% \end{figure}

\begin{figure}[t]
    \centering
    \includegraphics[width=0.98\linewidth]{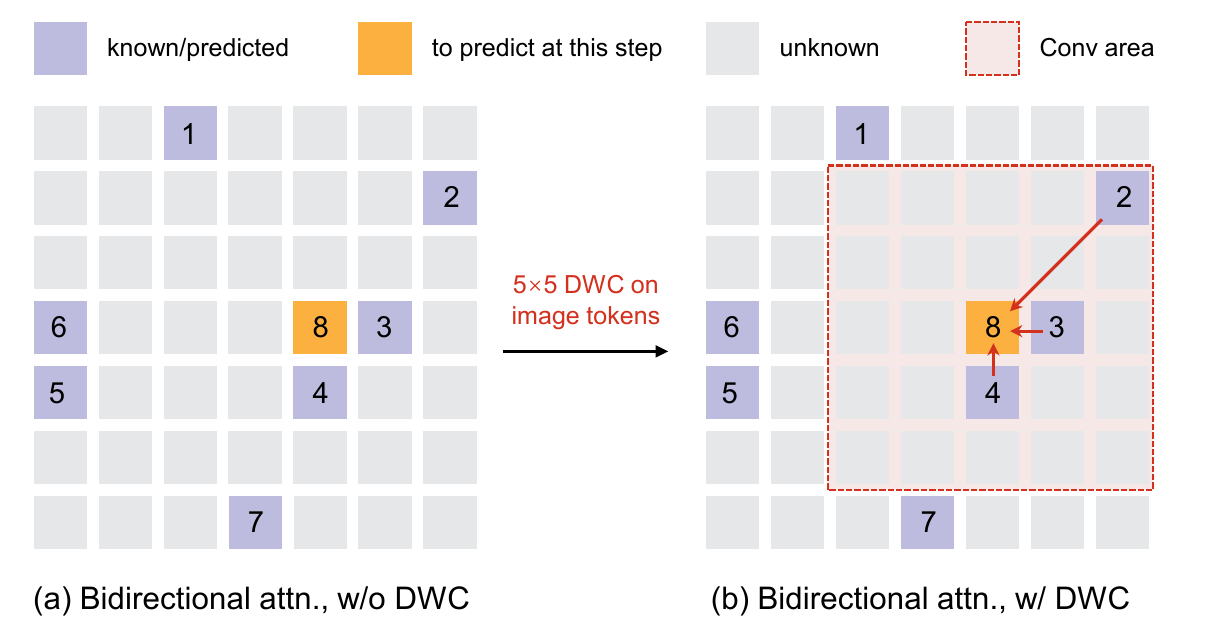}
    \caption{
    \textbf{DWC helps locality.} 
(a) A random-order autoregressive model with bidirectional attention predicts next tokens based on the predicted tokens. 
When the target token (\eg, the 8th) is surrounded by predicted tokens (\eg, the 3rd), the model faces challenges due to the limited local modeling capacity. 
(b) DWC module gathers information from nearby known tokens when predicting the current token, thereby facilitating linear attention.
    }
    \vspace{-1em}
    \label{fig:dwc_linear}
\end{figure}

% ************************************************
% Beyond scaling, we further draw inspiration from gating mechanisms in autoregressive language models to study memory control in our context.
% \textbf{Q3} (\textit{memory management}): Linear attention has a fixed memory size to process image tokens without clear differentiation. As tokens vary in semantics, long sequences may cause memory collisions~\citep{yang2024gated, schlag2021linear}, hindering the focus on informative tokens.
% ************************************************

% ************************************************
% Then, we introduce the KV Gate as a remedy for memory collisions and empirically evaluate several design choices (Sec.~\ref{sec:method.kvgate}, for \textbf{Q3}). 
% ************************************************

% \subsection{Scaling Behavior of Linear Attention: An Empirical Study} 
% \label{sec:method.scaling}

\subsection{Normalization Paradigm: Division \vs Subtraction}
\label{sec:scaling:normalization}

We empirically compare the scaling behaviors \wrt parameter counts of division-based and subtraction-based normalization linear attention (discussed in Sec.~\ref{sec:preliminary.la}) in the context of autoregressive generative modeling, even though both forms have already been studied in diffusion modeling using DiTs~\citep{peebles2023scalable} or image classification via ViTs~\cite{dosovitskiy2020image, touvron2021training}. We consider this comparison to be necessary. Unlike DiTs or ViTs, which generate all tokens in parallel, MAR models generate tokens sequentially in a step-by-step manner. At each step, they rely on previously predicted tokens to produce the next set. In the early stages of inference, only a few predicted tokens are available. Therefore, effectively exploiting the semantic context is vital for generation.

% Division-based normalization form has been shown effective in diffusion models~\citep{pu2024efficient, xie2024sana, xie2025sana, wang2025lit}, but it often comes with semantic confusion~\citep{han2024bridging}. 
% Subtraction-based normalization form endows the attention mechanism with an injective property, thereby avoiding this issue. 
% Both forms have been validated in image perception task as efficient alternatives to full attention~\citep{cai2023efficientvit, han2023flatten, guo2024slab}. But to our knowledge, it remains unclear which approach is more advantageous in autoregressive image generation task. In this work, we conduct a systematic empirical study of their scaling problem. 

\subsection{Locality Augmentation in Linear Attention}
\label{sec:scaling:dwc}

A key issue of linear attention is its capacity for local modeling~\citep{han2023flatten, han2024bridging}. Unlike softmax attention, it does not apply the \texttt{softmax} operation. 
Note that \texttt{Softmax} may account for the difference in locality modeling between softmax and linear attention.

\begin{equation}
\small
\begin{aligned}
H &=
\begin{bmatrix}
5.0 & 1.0 & 0.5 \\
0.5 & 4.0 & 1.0 \\
1.0 & 0.5 & 3.5
\end{bmatrix}, 
\\
A^{(\texttt{\texttt{f}})} &=
\begin{bmatrix}
0.9714 & 0.0178 & 0.0108 \\
0.0280 & 0.9259 & 0.0461 \\
0.0725 & 0.0440 & 0.8835
\end{bmatrix}\approx\begin{bmatrix}
1 & 0 & 0 \\
0 & 1 & 0 \\
0 & 0 & 1
\end{bmatrix}.
\label{softmax_example}
\end{aligned}
\end{equation}

% \begin{equation}
% \begin{aligned}
% \small
% H =
% \begin{bmatrix}
% 5.0 & 1.0 & 0.5 \\
% 0.5 & 4.0 & 1.0 \\
% 1.0 & 0.5 & 3.5
% \end{bmatrix}, 
% A^{(\texttt{\texttt{f}})}=\texttt{softmax}(H) =
% \begin{bmatrix}
% 0.9714 & 0.0178 & 0.0108 \\
% 0.0280 & 0.9259 & 0.0461 \\
% 0.0725 & 0.0440 & 0.8835
% \end{bmatrix}\approx\begin{bmatrix}
% 1 & 0 & 0 \\
% 0 & 1 & 0 \\
% 0 & 0 & 1
% \end{bmatrix}.
% \label{softmax_example}
% \end{aligned}
% \end{equation}

We provide a numerical example to clarify this effect. Consider a matrix $H$ (Eq.~\ref{softmax_example}). After applying \texttt{softmax}, the results $A^{(\texttt{f})}=\texttt{softmax}(H)$ becomes nearly identical to an identity matrix, implying that each token primarily attends to itself—representing the limiting case of local modeling. In this case, although $H$ may differ from the identity matrix, \texttt{softmax} helps sharpening the attention distribution, an essential operation that linear attention lacks. Thus, linear attention may face challenges in learning local relations.

\paragraph{Does scaling MAR models benefit from locality augmentation? A conceptual discussion.} 
As shown in Fig.~\ref{fig:dwc_linear}-(a), MAR models generate tokens set by set during inference. When the tokens to be predicted are adjacent to already known tokens, the latter provide valuable context, as neighboring tokens typically exhibit semantic correlations. 
We argue that, in this context, locality modeling is a crucial property in which linear attention is limited. 
As shown in Fig.~\ref{fig:dwc_linear}-(b), we apply a $5\times5$ \textit{depthwise convolution (DWC)}~\citep{chollet2017xception, howard2017mobilenets} module to potentially cooperate with linear attention, as its effectiveness has been extensively validated in both ViTs~\citep{han2023flatten, han2024bridging} and DiTs~\citep{wang2025lit}. 
% Fig.~\ref{fig:lam}-(b) illustrates an example. 
During inference, when predicting a token in an autoregressive step, DWC module incorporates neighboring predicted tokens (if available). 
If these known tokens are spatially close to the target token in 2D space, they tend to share similar semantics, providing useful cues for generation. 
On the other hand, pure ConvNets show limited scaling potential (\eg, ConvNeXt V2~\citep{liu2022convnet, woo2023convnext} at $\sim$650M parameters) compared with vision transformers (\eg, ViT-22B~\citep{dehghani2023scaling} at $\sim$22B parameters). As a result, we ablate the use of DWC as a design choice to examine its effect on scaling behavior \wrt parameter counts.

\paragraph{Implementation.} 
Unlike prior works~\citep{han2023flatten} that apply convolution to the value, we add DWC module to image features $I_{\text{img}}$ \textit{only} to the \model \textit{Decoder}—excluding the query features—and add its output to that of linear attention before the output projection of the attention: 

% Notably, we add DWC module to image features \textit{only}, excluding query features. 

\begin{equation} 
\small
\begin{aligned}
O^{(\texttt{d})}&=\texttt{LA}^{(\texttt{d})}(\left[ I_{\text{q}},I_{\text{img}} \right])+\texttt{DWC}(I_{\text{img}}),\\
O^{(\texttt{s})}&=\texttt{LA}^{(\texttt{s})}(\left[ I_{\text{q}},I_{\text{img}} \right])+\texttt{DWC}(I_{\text{img}}),
\label{eq:lam_equation}
\end{aligned}
\end{equation}

where $\texttt{LA}^{(\texttt{d})}$ and $\texttt{LA}^{(\texttt{s})}$ denote linear attention with division-based normalization and subtraction-based normalization, respectively.
The reasons are twofold. First, query features $I_{\text{q}}$ already incorporate textual conditioning, rendering the 2D inductive bias of DWC unnecessary. Second, \model encoder inputs consist only of randomly predicted image tokens, where reshaping them into a 2D layout produces neighborhoods that are not equivalent to those formed by the full set of image features, limiting the benefit of DWC at this stage.  
% With $L_{\texttt{D}}$ Decoder layers, DWC adds only $k \times k \times D \times L_{\texttt{D}} \approx 0.31$M parameters, negligible relative to the model size.
With $L_{\texttt{D}}$ decoder layers, DWC introduces only $k \times k \times D \times L_{\texttt{D}} \approx 0.31$M parameters, which is negligible compared with the overall model size of about 0.4B parameters.

\subsection{Scaling Behaviors}

\paragraph{Experimental setup.}
% In this section, we examine the impact of the linear attention paradigm and LAM on the scaling performance of \textit{linear autoregressive image generative models}. 
Based on the two design choices discussed above, we have four distinct settings. 
For each setting, we train models of three sizes, with $\sim$0.4B, $\sim$0.6B, and $\sim$1.4B parameters, respectively. 
The evaluation is conducted on ImageNet~\citep{deng2009imagenet} class-conditional image generation at 256$\times$256 resolution. 
All models are trained for 200K iterations on 32 A100 (40GB) GPUs with a learning rate of $8\times10^{-4}$. Inference is performed with BFloat16 precision. We report performance without classifier-free guidance (CFG)~\citep{ho2022classifier}, including FID-50K~\citep{heusel2017gans}, sFID~\citep{nash2021generating}, Inception Score~\citep{salimans2016improved}, and Precision/Recall~\citep{kynkaanniemi2019improved}. Detailed model configuration, hyper-parameters and results are presented in Appendix~\ref{sec:appendix.configuration},~\ref{sec:appendix.hyper-parameter}, and~\ref{sec:appendix.scaling-results}, respectively.

\begin{figure*}[t]
\centering
\includegraphics[width=0.92\textwidth]{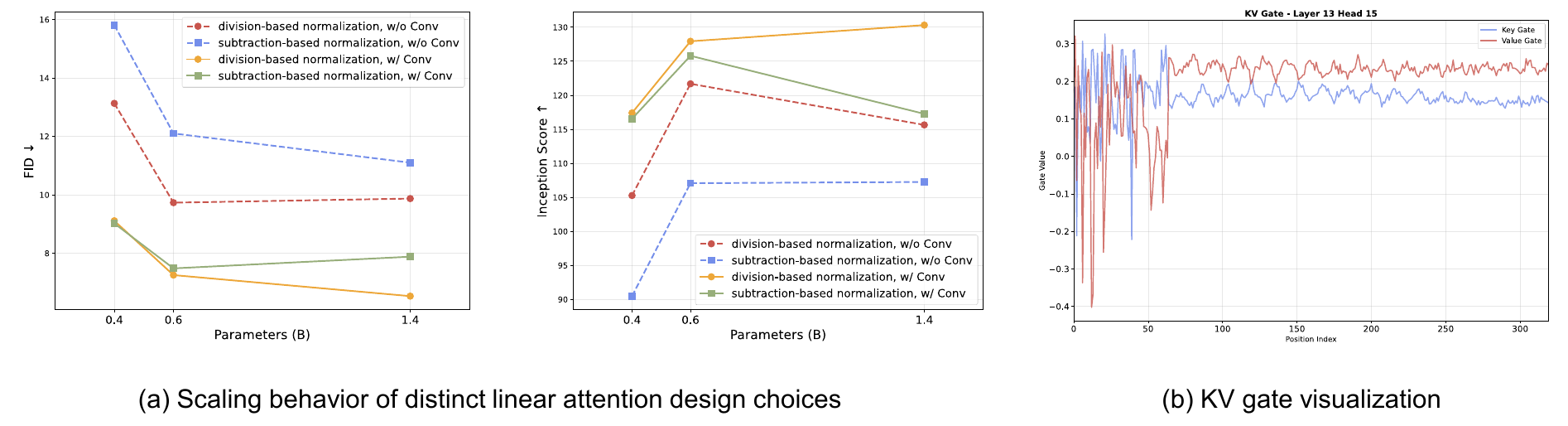}
\caption{
\textbf{Scaling behavior and KV gate results.} Fig.~(a) describes the class-conditional image generation results on the ImageNet 256$\times$256 benchmark using FID ($\downarrow$) and IS ($\uparrow$). Division-based linear attention with LAM achieves the best scaling performance. Detailed results are provided in Appendix~\ref{sec:appendix.scaling-results}. Fig.~(b) presents the learned \textit{KV gate} of a 256px text-to-image \model model. 
}
\vspace{-1.5em}
\label{fig:scaling_behavior}
\end{figure*}

\paragraph{Division-based normalization empirically scales better than subtraction-based normalization.}
Fig.~\ref{fig:scaling_behavior}-(a) illustrates the scaling performance of the four configurations, reported in terms of FID and IS. We observe that, regardless of the DWC module, division-based linear attention consistently outperforms subtraction-based at the huge scale ($\sim$1.4B). We hypothesize two possible reasons: (1) semantic confusion~\citep{han2024bridging} may be absent in autoregressive image generation, or its impact is less pronounced than in vision perception models; and (2) at early generation steps, when few tokens are predicted, masked tokens may interfere more with subtraction-based normalization.

\paragraph{DWC module tends to improve both linear attention variants across model sizes.}
From Fig.~\ref{fig:scaling_behavior}-(b), we further observe that for both forms of linear attention, adding DWC to image features consistently improves performance (\ie, FID and IS) across model sizes. We attribute this to the limited locality of linear attention used, which remains a bottleneck in autoregressive image generation. Introducing an appropriate inductive bias, \eg, depthwise convolution, appears beneficial for parameter scaling. Developing effective ways to enhance locality remains an interesting direction for future work. 
From now on, we will use division-based linear attention with DWC as our basic design choice.

\section{KV Gate For Flexible Memory Management} 
\label{sec:gating}

Gating mechanism~\cite{yang2023gated, yang2024gated, sun2023retentive, sun2024you, dao2024transformers, schlag2021linear}, a common practice in autoregressive language models, help flexibly manage memory. In \textit{causal} linear attention, they are typically implemented as data-dependent decay terms to selectively ``forget" past information. 
Gating factor not only preserves efficiency but also shows potential in language modeling, \etc~\citep{yang2024gated}.
However, their use in settings that require \textit{bidirectional} attention (\eg, autoregressive image models) remains underexplored. 
We believe the key lies in the difference between causal and bidirectional mechanisms. 
Denote the memory as $M$. In causal attention, the forget gate $\alpha$ is used to ``erase'' past information, \ie, $M_t=\alpha_tM_{t-1}+K_t^\top V_t$ (assume $\phi(\cdot)$ is identity function). 
However, in bidirectional attention, there is no causal mask and thus no notion of a strict ``past'': tokens at any positions in the sequence can attend to each other. 
Therefore, we argue that the memories $M_i$ and $M_j$ at arbitrary positions $i$ and $j$ should not have a forget--retain relationship. 
Instead, they should be assigned different importance weights.  

To this end, we propose a simple yet effective method, called \textit{KV gate}, to equip bidirectional linear attention with gating. 
We express linear attention (Eq.~\ref{eq:division_based_linear}) as the following form:

% \vspace{-0.25em}

% \begin{equation}
% \begin{gathered}
% \small
% M=\sum_{j=1}^{N}{\phi(K_j)^\top V_j }=\sum_{j=1}^{N}{M_j},\\
% z=\sum_{m=1}^{N}{\phi(K_m)^\top},~~
% O^{(\texttt{d})}_i          
% = \frac{\phi(Q_i) M}{\phi(Q_i) z},
% \label{eq:kvgate_vanilla}
% \end{gathered}
% \end{equation} 

\begin{equation}
\small
\begin{gathered}
M=\sum_{j=1}^{N}{\phi(K_j)^\top V_j },
z=\sum_{m=1}^{N}{\phi(K_m)^\top},
O^{(\texttt{d})}_i          
= \frac{\phi(Q_i) M}{\phi(Q_i) z},
\label{eq:kvgate_vanilla}
\end{gathered}
\end{equation} 
% \vspace{-0.25em}

where memory $M\in\mathbb{R}^{D\times D}$ represents the \textit{equally weighted sum} of token-wise memories $M_i$, while $z\in\mathbb{R}^{D\times 1}$ contributes to the normalization term ${\gamma^{(\texttt{d})}_i=\phi(Q_i) z}$.

% Modern linear LLMs~\citep{yang2023gated, yang2024gated, sun2023retentive, sun2024you, dao2024transformers, schlag2021linear} adopt gated linear attention (GLA), which employs a forget gate to discard past information selectively.
% GLA not only preserves efficiency but also shows remarkable potential in language modeling and commonsense reasoning tasks, \etc~\citep{yang2024gated}. 
% We argue that in autoregressive image generation, memory corresponding to different tokens should be treated suitably, introduced next. 

% \paragraph{How to effectively achieve flexible memory arrangement?} 

\textit{In a nutshell, we use a learnable K gate to scale both $M$ and $z$, and another learnable V gate to scale $M$ only. This method, dubbed \textit{KV gate}, allows effective and flexible memory arrangement.} Division-based normalization linear attention (Eq.~\ref{eq:kvgate_vanilla}) using the \textit{KV gate} can be formulated as follows: 

% \vspace{-0.25em}

\begin{equation}
\small
\begin{gathered}
{\tilde{K}_j}={g_j^{(k)}}\phi(K_j), ~~
{\tilde{V}_j}={g_j^{(v)}}V_j,~~\textit{for}~~j\in \left[1,N\right]
\\
M=\sum_{j=1}^{N}{ 
{\tilde{K}_j}^\top{\tilde{V}_j} 
}=
\sum_{j=1}^{N}{{g_j^{(k)}}{g_j^{(v)}}M_j},\\
z=\sum_{m=1}^{N}{{\tilde{K}_m}^\top},~~
O^{(\texttt{d})}_i          
= \frac{\phi(Q_i) M}{\phi(Q_i) z},
\label{eq:kvgate_kv}
\end{gathered}
\end{equation} 

% \vspace{-0.25em}

where the \textit{K gate} ${g^{(k)}}$ denotes the scaling coefficient applied to $\phi(K)$ when computing both $M$ and $z$, while the \textit{V gate} ${g^{(v)}}$ serves as an auxiliary part that adjusts $M$ only. 

\begin{table}[t]
    \centering
    \caption{\textbf{Ablation study of KV gate.} ImageNet 256$\times$256 results (w/o CFG) are reported.
    All models are trained for 200K iterations. Head-wise KV gate are chosen. HW: head-wise. HS: head-shared. 
    } 
    \vspace{0em}
    \label{tab:kv_gate_result}
    \scalebox{0.74}{ 
    \begin{tabular}{lcccccccc}
        \toprule
        Gate &  Key & Value & $z^{(\texttt{d})}$ & FID~$\downarrow$ & sFID~$\downarrow$ & IS~$\uparrow$ & Pre.~$\uparrow$ & Rec.~$\uparrow$ \\
        \midrule
        None &    &  &  & 9.11 & 5.89 & 117.40 & 0.69 & 0.61 \\
        \midrule
        HW  & \checkmark & \checkmark &  & \textbf{8.72} & 5.64 & \textbf{120.34} & 0.69 & 0.61 \\
        HW  & \checkmark &  &  & 8.80 & 5.69 & 119.20 & 0.70 & 0.62 \\
        HW  &  & \checkmark &  & 9.06 & 5.60 & 115.90 & 0.69 & 0.61 \\
        HW  & \checkmark & \checkmark & \checkmark & 9.22 & 5.89 & 115.41 & 0.69 & 0.61 \\
        \midrule
        HS  & \checkmark & \checkmark &  & \textbf{8.72} & 5.70 & 120.24 & 0.69 & 0.62 \\
        \bottomrule
        \end{tabular}
    }
    \vspace{-1.5em}
\end{table}

KV gate is applied in all linear attention in the \textit{Decoder} of our model. 
Besides the KV gate, we also design three variants as ablations, \ie, \textit{K gate only}, \textit{V gate only}, and a variant with an \textit{extra gate} applied to $z$ (see Appendix E for details of the four modes). Ablation study is conducted on ImageNet class-conditional image generation task using a $\sim$0.4B model. 

As reported in Tab.~\ref{tab:kv_gate_result}, KV gate consistently improves FID, IS, and sFID compared to division-based normalization linear attention baseline. 
We present two thought-provoking findings. 
\textbf{(1)} Using only V gate or applying an extra gate to $z$ severely degrades FID and IS. 
\textbf{(2)} Using only K gate slightly affects FID and IS.
This indicates that the K gate and V gate exhibit a synergistic effect rather than being mutually exclusive. Moreover, we suggest leveraging the K gate to scale $z$, instead of introducing an additional set of learnable coefficients.
With $h$ attention heads, the KV gate introduces only $2\times h\times N \times L_{\texttt{D}}\approx 0.12$M parameters for a sequence length of $N=320$, which is negligible compared to the $\sim$0.4B parameters of the model.

% We also visualize the \textit{KV} gates to illustrate the specific patterns they have learned. As shown in Fig.~\ref{fig:scaling_behavior}-(b), for the first 64 query tokens, the values of the \textit{KV} gates exhibit fluctuations, while for the subsequent 256 image tokens, they display a clear periodic pattern.  
% This indicates that when flexibly managing memory $M$, the scaling of $z$ is also important: for the $j$-th token, both $M$ and $z$ should be computed with the same scaling coefficient applied to its key $\phi(K_j)$. We attribute this to the fact that $z$ contributes to the normalization term; therefore, inconsistent or omitted scaling may lead to unstable training and degraded model performance. 

% \subsection{Analysis of KV Gate}

\paragraph{Should the KV gate be head-wise or head-shared?}
We investigate whether the KV gates across different attention heads should share parameters.
As shown in Tab.~\ref{tab:kv_gate_result}, using a head-specific KV gate slightly improves IS and sFID. Given its negligible parameter overhead, we adopt the head-wise design. 
We kindly note that~\citep{lin2025forgetting, qiu2025gated} investigate incorporating gating mechanisms into softmax attention. 
% We leave for future work a systematic study of the similarities and differences in how gating influences linear attention \vs full attention. 

\paragraph{What pattern has the KV gate learned?} 
As shown in Fig.~\ref{fig:scaling_behavior}-(b), we visualize the KV gate of \model-H trained for 500K iterations at 256px resolution on text-to-image generation task to illustrate specific patterns they have learned.
For the first 64 query tokens, the values of the KV gate exhibit fluctuations, while for the subsequent 256 image tokens, they are more stable and display a clear periodic pattern. 
Notably, K gate peak when V gate dip, indicating complementary roles in memory management in the linear attention. Additional results (provided in Appendix~\ref{sec:appendix.kvgate-results}) show that, although we do not explicitly restrict the KV gate values, they generally remain within the interval $(0,1)$ and display diverse patterns across different layers and heads. See Appendix~\ref{sec:appendix.kvgate-results} for detailed results.

\paragraph{Difference between KV gate and the forget gate.}
% Gated linear attention~\citep{sun2023retentive, yang2023gated, yang2024gated} in LLMs often introduce a forget gate as a decay term to forget past information, \ie, $M_t=\alpha_tM_{t-1}+K_t^\top V_t$ (assume $\phi(\cdot)$ is the identity).
Forget gate in GLA and our KV gate both modulates memory but differs in three ways. 
\textbf{(1)} \textit{Recurrence vs.\ parallelism:} Forget gate is recursive---its decay factor for $M_j$ is $\prod_{s=j+1}^{t}\alpha_s$. In contrast, our KV gate computes the factor ${g_j^{(k)}}{g_j^{(v)}}$ once without recurrence. 
\textbf{(2)} \textit{Data dependence:} Forget gate is typically data-dependent, with $\alpha_t$ projected from the current input token~\citep{yang2023gated}. Differently, KV gate is data-independent. We find that simple learnable parameters is sufficient to improve performance without linear projection. 
\textbf{(3)} \textit{Range constraint:} Forget gate~\citep{sun2024you} restricts $\alpha_t$ to $(0,1)$ due to the \texttt{sigmoid} function, whereas KV gate allows ${g_j^{(k)}}$, ${g_j^{(v)}}$ to be negative, supporting flexible learning.

% 实际可视化的是 1536 channel，训练500K iteration的T2I模型权重的KV Gate结果。
\section{Experiments}
\label{sec:experiments}

\subsection{Class-conditional Image Generation}
\label{sec:experiments.c2i}

\paragraph{Training details.}

As an system-level validation of our method, we conduct class-conditional image generation experiments on the ImageNet 256$\times$256 benchmark using \model-H (1.4B). 
Using a learning rate of $8\times10^{-4}$ and a batch size of $768$, we train the model for a total of 1.2M iterations. We set the model EMA~\citep{polyak1992acceleration} to $0.99$ and the weight decay to $0.02$.
Following NOVA, we adopt a 6-layer flow matching MLP as the denoising network, without relying on advanced training approaches, \eg, REPA~\citep{yu2024representation}. 
Sampling during inference is done in BFloat16 precision. The autoregressive and diffusion steps are set at 64 and 25, respectively.  We report results with CFG scale of 1.0 and 2.4. 

\paragraph{Results.}
In Tab.~\ref{tab:c2i_main}, we present a system-level evaluation of \model against other frontier models.  
Compared with the linear diffusion model (\eg, LiT~\cite{wang2025lit}) and the full attention autoregressive model (\eg, MAR), \model—based on a linear autoregressive architecture—delivers competitive performance. Notably, LINA-H achieves an FID of 2.18, demonstrating that the linear attention designed in this work is well suited for autoregressive image modeling. 
In addition, LINA attains superior validated performance than DiM, an efficient state space model~\citep{gu2023mamba, dao2024transformers} based diffusion model. As a result, we turn to validate \model on text-to-image benchmarks.

\begin{table}[t]
\centering
\caption{\textbf{Class-conditional ImageNet 256$\times$256 results.}
}
\vspace{-0.5em}
\label{tab:c2i_main}
\scalebox{0.69}{
\begin{tabular}{lcccc}
\toprule
\textbf{Model} & \textbf{FID}$~\downarrow$ & \textbf{IS}$~\uparrow$ & \textbf{Precision}$~\uparrow$ & \textbf{Recall}$~\uparrow$ \\
\midrule
\multicolumn{5}{c}{\textbf{\textit{Diffusion models}}} \\
\addlinespace[1pt]
ADM~\citep{dhariwal2021diffusion} & 4.59 & 186.70 & 0.82 & 0.52 \\
CDM~\citep{ho2022cascaded} & 4.88 & 158.71 & - & - \\
LDM-4~\citep{rombach2022high} & 3.60 & 247.67 & 0.87 & 0.48 \\
U-ViT-H/2-G~\citep{bao2022all} & 2.29 & 263.9 & 0.82 & 0.57 \\
DiT-XL/2~\citep{peebles2023scalable} & 2.27 & 278.24 & 0.83 & 0.57 \\
LiT-XL/2~\citep{wang2025lit} & 2.32 & 265.20 & 0.824 & 0.574 \\
DiffuSSM-XL-G~\citep{yan2024diffusion} & 2.28 & 259.13 & 0.86 & 0.56 \\
DiM-L~\citep{teng2024dim} & 2.64 & - & - & - \\
DiM-H~\citep{teng2024dim} & 2.21 & - & - & - \\
DiG-XL/2-G~\cite{zhu2024dig} & 2.07 & 278.95 & 0.82 & 0.60 \\
SiT-XL~\citep{ma2024sit} & 2.06 & 277.50 & 0.83 & 0.59 \\
Mediator~\cite{pu2024efficient} & \underline{2.01} & 271.04 & 0.82 & 0.60 \\
\midrule
\midrule
\multicolumn{5}{c}{\textbf{\textit{Autoregressive models}}} \\
\addlinespace[1pt]
Mask-GIT~\citep{chang2022maskgit} &  6.18 & 182.1 & - & - \\
MAGVIT-v2~\citep{yu2023language} & \textbf{1.78} & \textbf{319.4} & - & - \\
MAR-B~\citep{li2024autoregressive} & 2.31 & 281.7 & 0.82 & 0.57 \\
MAR-L & \textbf{1.78} & \underline{296.0} & 0.81 & 0.60 \\
\midrule
LINA-H~(cfg=1.0) & 4.49 & 162.64 & 0.74 & 0.62 \\
LINA-H~(cfg=2.4) & 2.18 & 275.73 & 0.81 & 0.58 \\
\bottomrule
\end{tabular}
}
\vspace{-1.5em}
\end{table}

\subsection{Text-to-image Generation}
\label{sec:experiments.t2i}
% stage1: 565K iterations 

\paragraph{Training details.}
Our training pipeline consists of three stages. Following LiT, we initialize stage 1 with the 1024px NOVA pretrained weights, excluding the linear attention. The three stages are trained on 256px, 512px, and 1024px data, respectively. Training is conducted on 48 A100 (40GB) GPUs with batch sizes of 768, 192, and 48. Stages 2 and 3 run for 600K and 700K iterations. We set 128 autoregressive steps and 25 diffusion steps, with a CFG scale of 7.0 during sampling. Details are provided in Appendix~\ref{sec:appendix.t2isetting}.

\paragraph{Results.}

As shown in Tab.~\ref{tab:t2i_main}, 
we compare \model with advanced text-to-image architectures on the GenEval benchmark~\citep{ghosh2023geneval}. Without prompt engineering, the 1.4B \model outperforms the 1.6B SANA, an advanced linear DiT baseline. Moreover, compared with autoregressive models using full attention, \model achieves performance on par with the 10.5B Fluid. These results demonstrate that the proposed linear attention integrates well with autoregressive architectures and exhibits strong text-image alignment, providing a clear and reliable baseline. 
Fig. 1 shows 1024px samples generated by \model, where image fidelity and fine textures are well-preserved across long sequences.

\begin{table*}[t]
\centering
\caption{\textbf{Comparison of GenEval results.}
Rewriter refers to a prompt engineering method~\citep{deng2024autoregressive}. Our \model, equipped with pure linear attention, rivals advanced T2I frameworks. Best results are in \textbf{bold}; second best are \underline{underlined}.   
}
\vspace{-0.5em}
\label{tab:t2i_main}
\scalebox{0.8}{
\begin{tabular}{lcccccccc}
\toprule
\textbf{Model} & \textbf{Params.} & \textbf{Overall} & \textbf{Single} & \textbf{Two} & \textbf{Counting} & \textbf{Colors} & \textbf{Position} & \textbf{ColorAttr} \\
\midrule
\multicolumn{8}{c}{\textbf{\textit{Diffusion models}}} \\
\addlinespace[1pt]
PixArt-$\alpha$~\citep{chen2024pixart1} & 0.6B & 0.48 & 0.98 & 0.50 & 0.44 & 0.80 & 0.08 & 0.07 \\
LiT~(1024$\times$1024)~\citep{wang2025lit} & 0.6B & 0.48 & 0.98 & 0.50 & 0.40 & 0.77 & 0.11 & 0.12 \\
LiT~(512$\times$512) & 0.6B & 0.47 & 0.97 & 0.43 & 0.42 & 0.79 & 0.09 & 0.12 \\
DALL-E3~\citep{Dalle-3} & - & 0.67 & 0.96 & 0.87 & 0.47 & 0.83 & 0.43 & 0.45 \\
SDXL~\citep{podell2023sdxl} & 2.6B & 0.55 & 0.98 & 0.44 & 0.39 & 0.85 & 0.15 & 0.23 \\
SD3~\citep{esser2024scaling} & 2B & 0.62 & 0.98 & 0.74 & 0.63 & 0.67 & 0.34 & 0.36 \\
Playground v2.5~\citep{li2024playground} & 2.6B & 0.56 & - & - & - & - & - & - \\
Hunyuan-DiT~\citep{li2024hunyuan} & 1.5B & 0.63 & - & - & - & - & - & - \\
SANA~(1024$\times$1024)~\citep{xie2024sana} & 1.6B & 0.66 & - & - & - & - & - & - \\
SANA~(512$\times$512) & 1.6B & 0.66 & - & - & - & - & - & - \\
\midrule
\midrule
\multicolumn{8}{c}{\textbf{\textit{Autoregressive models}}} \\
\addlinespace[1pt]
LlamaGen~\citep{sun2024autoregressive} & 0.8B &  0.32 & 0.71 & 0.34 & 0.21 & 0.58 & 0.07 & 0.04 \\
Emu3 (+ Rewriter)~\citep{wang2024emu3}  & 8B & 0.66 & 0.99 & 0.81 & 0.42 & 0.80 & 0.49 & 0.45 \\
Show-o~\citep{xie2024show} & 1.3B & 0.53 & 0.95 & 0.52 & 0.49 & 0.82 & 0.11 & 0.28 \\
NOVA~(1024$\times$1024)~\citep{deng2024autoregressive} & 1.4B & 0.71 & 0.99 & 0.91 & 0.62 & 0.85 & 0.33 & 0.56 \\
NOVA~(512$\times$512) (+ Rewriter)  & 0.6B & \textbf{0.75} & 0.98 & 0.88 & 0.62 & 0.82 & 0.62 & 0.58 \\
Fluid~\citep{fan2024fluid} & 1.1B & 0.67 & 0.96 & 0.77 & 0.61 & 0.78 & 0.34 & 0.53 \\
Fluid~\citep{fan2024fluid} & 10.5B & 0.69 & 0.96 & 0.83 & 0.63 & 0.80 & 0.39 & 0.51 \\
\midrule
LINA-H~(1024$\times$1024) & 1.5B & 0.66 & 0.99 & 0.85 & 0.50 & 0.88 & 0.38 & 0.39 \\
\,\, + Rewriter & 1.5B & 0.72 & 0.99 & 0.84 & 0.54 & 0.85 & 0.56 & 0.53 \\
LINA-H~(512$\times$512) & 1.4B & 0.68 & 0.98 & 0.83 & 0.56 & 0.89 & 0.34 & 0.50 \\
\,\, + Rewriter & 1.4B & \underline{0.74} & 0.99 & 0.85 & 0.61 & 0.87 & 0.60 & 0.53 \\
\bottomrule
\end{tabular}
}
\vspace{-1em}
\end{table*}

\paragraph{FLOPs comparison: softmax attention \vs linear attention in \model.}
We report the FLOPs of \textit{a single} attention module under the following configuration: batch size of~1, sequence length of~5120, hidden dimension of~1536, and 16 attention heads. This setup matches how \model-H operates at a resolution of~1024px.
The linear attention module we evaluate adopts \textit{division-based normalization} in the \model \textit{Decoder} and integrates both the \textit{DWC module} and the \textit{KV gate} proposed in this work. FLOPs are measured using the fvcore~\citep{fvcore} library.

\begin{figure}[t]
\centering
\includegraphics[width=0.73\linewidth]{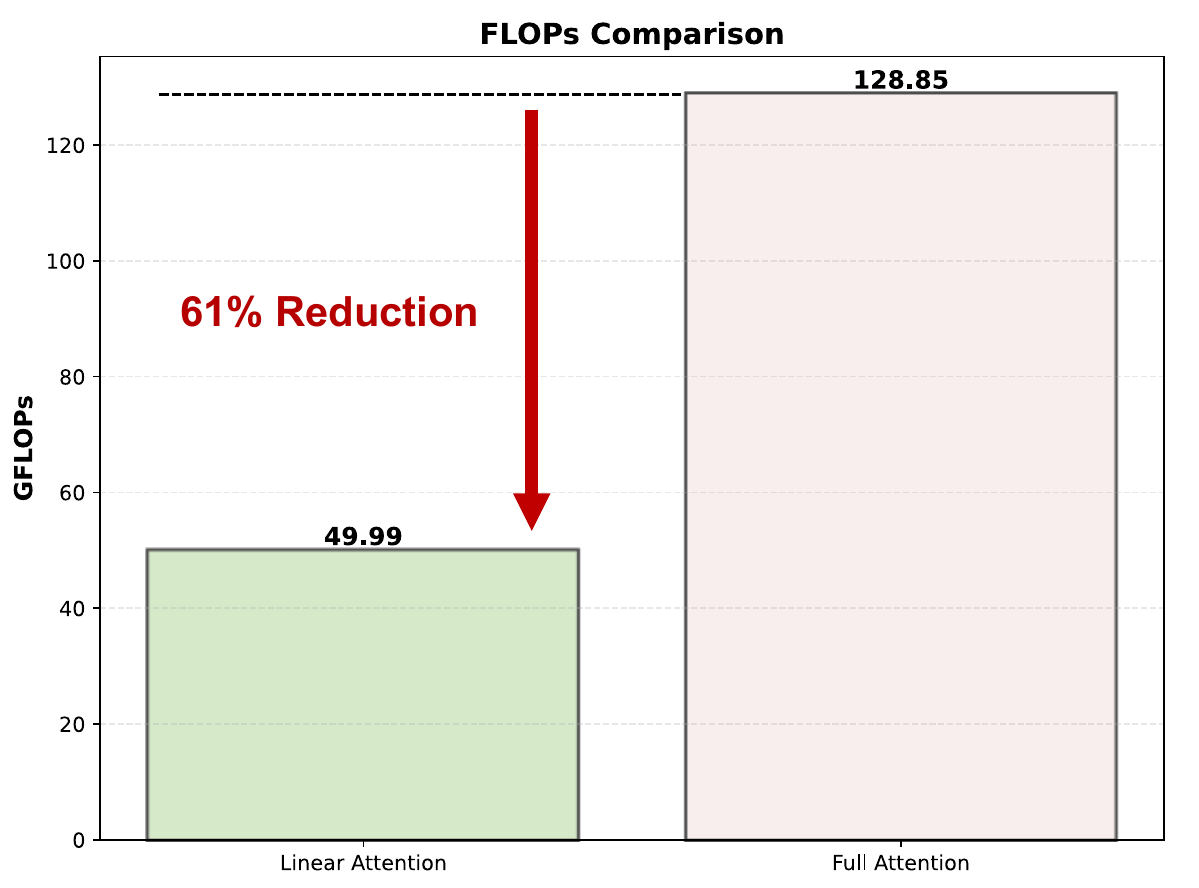}
\vspace{-0.5em}
\caption{
\textbf{FLOPs comparison: \textit{a single} module of linear attention \vs softmax attention.} We use a batch size of 1, a sequence length of 5120, a hidden dimension of 1536, and 16 attention heads. Such configuration corresponds to how \model-H operates at 1024px. Linear attention applies division-based normalization and incorporates both the \textit{DWC module} and the \textit{KV gate}. 
Compared with softmax attention, linear attention reduces FLOPs by $\sim$61\%, showing computation efficiency.
}
\vspace{-1.6em}
\label{fig:flops_comparison}
\end{figure}

The results are presented in Fig.~\ref{fig:flops_comparison}. Softmax attention requires approximately $\sim$129~GFLOPs, whereas the linear attention requires only $\sim$50~GFLOPs. This corresponds to a reduction of about $\sim$61\% in FLOPs, highlighting the efficiency of our \model. Importantly, despite this substantial reduction in FLOPs, the T2I performance of \model remains competitive with softmax attention-based NOVA.
Note that \textit{DWC module} and \textit{KV gate} are used only in the \model \textit{Decoder}, and are not applied in the \textit{Encoder} or the \textit{Connector}.

% \begin{figure}[h]
% \centering
% \begin{minipage}{0.44\linewidth}{
% \vspace{-1.5em}
% \includegraphics[width=0.99\textwidth]{figure/flops_comparison.pdf}
% }\end{minipage}
% % \hfill
% \begin{minipage}{0.53\linewidth}{
% \caption{
% \color{blue}{\textbf{FLOPs comparison results.} 
% We compare \textit{a single} module of linear attention and full attention under the following setting: batch size 1, sequence length 5120, hidden dimension 1536, and 16 heads. This configuration corresponds to how \model-H operates at 1024px resolution. The linear attention variant applies division-based normalization and incorporates both the \textit{DWC module} and the \textit{KV gate}, explored heavily in this work. 
% Compared with the full attention, \textit{a single} linear attention module reduces FLOPs by $\sim$61\%, showing computation efficiency.  }
% }
% \label{fig:flops_comparison}
% }\end{minipage}
% \vspace{-1em}
% \end{figure}

\paragraph{Latency comparison.}
Tab.~\ref{tab:latency} reports the pipeline latency comparison between NOVA and \model when generating 1024px images. \model shares the same macro architecture as NOVA, but completely replaces NOVA's softmax attention with linear attention. For each run, we compute the average latency of generating 10 images, and the reported latency is the mean over three such runs. All experiments are conducted on a NVIDIA A800 GPU with a batch size of 1. 
Since our work represents an early exploration of linear MAR models with an emphasis on sample quality, we do not incorporate additional acceleration techniques, \eg, Triton~\citep{tillet2019triton}.
Even without such optimizations, LINA attains latency on par with softmax attention using FlashAttention~\citep{dao2022flashattention}. We attribute this parity to the inherent computational advantage of linear attention in processing long sequences.

% Our design serves as a building block for linear MAR, readily complementing future acceleration methods. 

% \begin{wraptable}{r}{0.5\columnwidth}
%     \centering
%     \vspace{-1.1em}
%     \caption{\textbf{Latency comparison results.} \model competes FlashAttention in 1024px generation.}
%     \vspace{-0.6em}
%     \label{tab:latency}
%     \scalebox{0.7}{
%     \begin{tabular}{lccccc}
%     \toprule
%     \textbf{Model} & \textbf{Params.} & \textbf{Res.} & \textbf{Type} & \textbf{Acceleration} & \textbf{Latency} \\
%     \midrule
%     NOVA & 1.4B & 1024px & Full & FlashAttn & 20.0s \\
%     \model & 1.5B & 1024px & Linear & - & 22.0s  \\
%     \bottomrule
%     \end{tabular}
%     }
%     \vspace{-1.2em}
% \end{wraptable}

\begin{table}[h]
\centering
\caption{\textbf{Latency comparison results.} \model competes FlashAttention in 1024px generation.
}
\vspace{-0.5em}
\label{tab:latency}
\scalebox{0.78}{
\begin{tabular}{lccccc}
\toprule
\textbf{Model} & \textbf{Params.} & \textbf{Res.} & \textbf{Type} & \textbf{Acceleration} & \textbf{Latency} \\
\midrule
NOVA & 1.4B & 1024px & Softmax & FlashAttn & 20.0s \\
\model & 1.5B & 1024px & Linear & - & 22.0s  \\
\bottomrule
\end{tabular}
}
\vspace{-0em}
\end{table}

% Tab.3 给出了NOVA和LINA模型在生成1024px图片时候的延时比较结果。NOVA模型与我们的模型共享相同的宏观架构，但是全部使用full attention。
% 每次生成过程，我们取生成10张图片的均值。我们取三次生成过程的平均值，作为报告的latency。实验在1 台 NVIDIA A800 GPU进行，使用batch size为1测试。
% 考虑到本文聚焦于线性自回归模型的生成质量，因此并未对其加速技术 (如Triton) 做深入探索。尽管如此，LINA模型，在延时上与使用了Flash attention的full attention相当。我们把这一结果归结为线性注意力在处理较长序列时天然的计算量优势。本文的设计可以作为线性自回归模型的building block，服务其他加速方法。

\section{Conclusion}
\label{sec:conclusion}

This paper takes a deep dive into how linear attention should be designed for autoregressive image generative model with continuous tokens. We recommend adopting division-based normalization and incorporating convolution to strengthen locality. Besides, we introduce the KV gate, a simple way that modulates key and value states to enable flexible memory management and, in turn, improve generation. Our final model, dubbed \model, is an linear autoregressive model that delivers competitive image generation performance. 
% We envision this work as a building block that paves the way for developing acceleration techniques in linear attention toward real-world applications, which we leave for future work.
\section*{Impact Statements}
\label{sec:impact}
This paper presents work whose goal is to advance the field of machine learning, with a focus on efficient autoregressive image generation. There are many potential societal consequences of generative models, none of which we feel must be specifically highlighted here beyond those already well studied in the literature.

\bibliography{example_paper}
\bibliographystyle{icml2026}

%%%%%%%%%%%%%%%%%%%%%%%%%%%%%%%%%%%%%%%%%%%%%%%%%%%%%%%%%%%%%%%%%%%%%%%%%%%%%%%
%%%%%%%%%%%%%%%%%%%%%%%%%%%%%%%%%%%%%%%%%%%%%%%%%%%%%%%%%%%%%%%%%%%%%%%%%%%%%%%
% APPENDIX
%%%%%%%%%%%%%%%%%%%%%%%%%%%%%%%%%%%%%%%%%%%%%%%%%%%%%%%%%%%%%%%%%%%%%%%%%%%%%%%
%%%%%%%%%%%%%%%%%%%%%%%%%%%%%%%%%%%%%%%%%%%%%%%%%%%%%%%%%%%%%%%%%%%%%%%%%%%%%%%
\newpage
\appendix
\onecolumn

% You can have as much text here as you want. The main body must be at most $8$
% pages long. For the final version, one more page can be added. If you want, you
% can use an appendix like this one.

% The $\mathtt{\backslash onecolumn}$ command above can be kept in place if you
% prefer a one-column appendix, or can be removed if you prefer a two-column
% appendix.  Apart from this possible change, the style (font size, spacing,
% margins, page numbering, etc.) should be kept the same as the main body.

% \section{The Use of Large Language Models (LLMs)}
% \label{sec:appendix.llmusage}

% To ensure accuracy in language and faithfully convey the intended ideas, this paper was prepared with the assistance of large language models (LLMs). While not all generated outputs were adopted, LLMs directly or indirectly contributed to the writing process. In particular, the LLM was used for: 
% 1) Refining language expression by making the author’s draft more concise, natural, and clear.
% 2) Assessing whether the text was idiomatic and precise, and suggesting potential revisions.
% 3) Checking for possible grammatical errors and offering corrections.
% 4) Assisting in parts of the experimental code.
% The author remains fully responsible for the authenticity and integrity of all content in this paper.

\section{Full Related Work}
\label{sec:appendix.related}

% \paragraph{Efficient image generation.}
Efficiency is a key concern for image generative models when dealing with long sequences.
Research on efficient diffusion models is relatively extensive. Some approaches focus on architectural modifications, especially efficient attention~\citep{katharopoulos2020transformers, choromanski2020rethinking, yang2023gated, cai2023efficientvit, han2023flatten}. For example, SANA~\citep{xie2024sana} studies the text encoder in text-to-image models, while LiT~\citep{wang2025lit} provides guidelines for converting a pretrained DiT into a linear DiT. DiG~\citep{zhu2024dig} and DiM~\citep{teng2024dim} explore applying gated linear attention and state space models~\citep{gu2023mamba, dao2024transformers}, respectively, to image generation. Other works pursue fewer or even single diffusion steps, such as DMD~\citep{yin2024one}, DMD2~\citep{yin2024improved}, CausVid~\citep{yin2025slow}, consistency models~\citep{song2023consistency, lu2024simplifying}, and SANA-Sprint~\citep{chen2025sana}.

For autoregressive models with continuous tokens, researchers have explored various directions to improve efficiency. 
Notably, DiSA~\citep{zhao2025disa} reduces the number of diffusion steps as the autoregressive process progresses. 
LazyMAR~\citep{yan2025lazymar} explores how to use feature caching to improve efficiency while maintaining performance.  
DC-AR~\citep{wu2025dc} studies the design and training of image tokenizer. 
ARFlow~\citep{hui2025arflow} enables flow-based image generation through hybrid linear attention.

\section{Inference Pipeline}
\label{sec:appendix.inference}
\begin{figure}[!ht]
    \centering
    \includegraphics[width=0.94\linewidth]{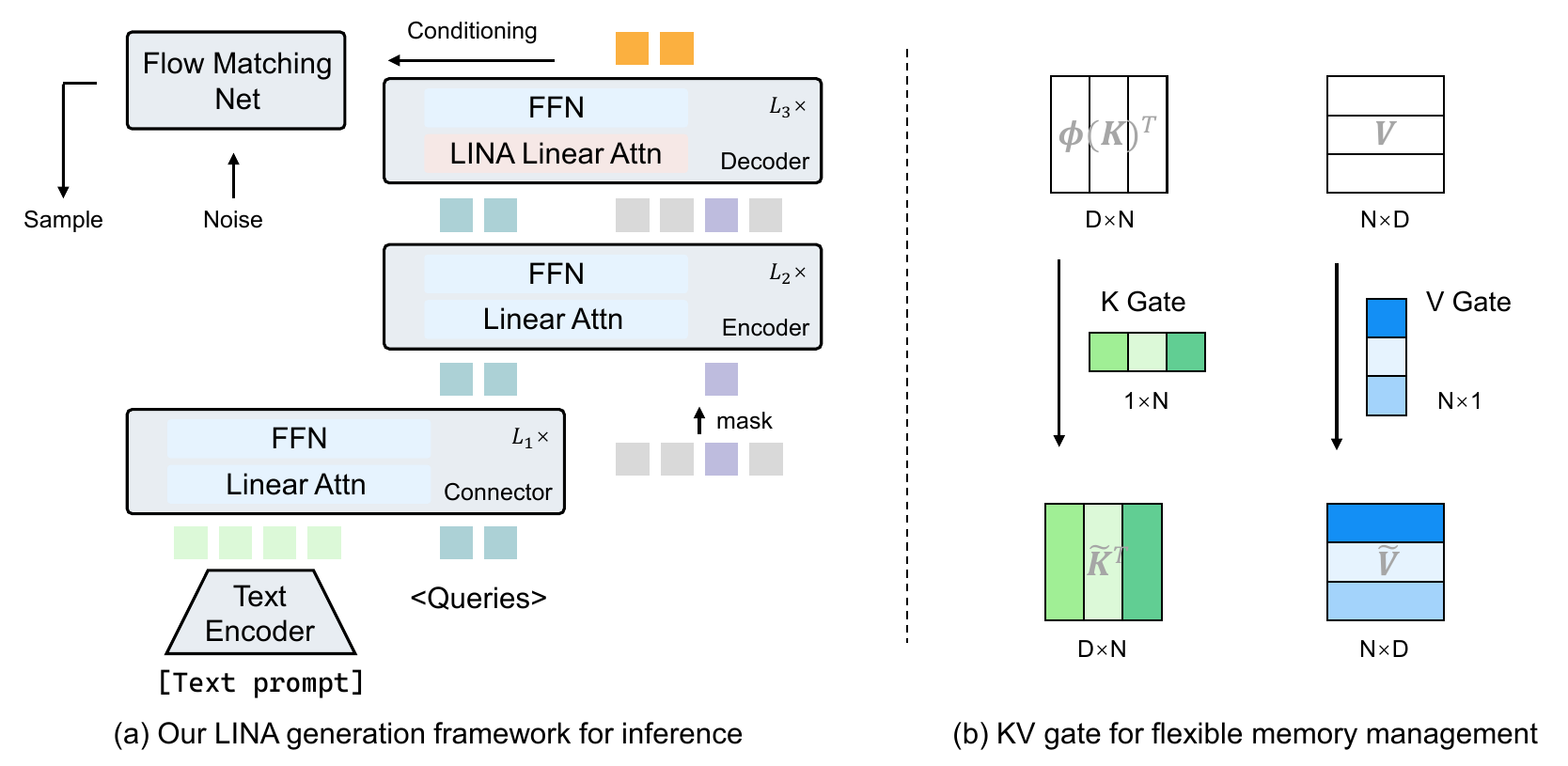}
    \caption{\textbf{Overview of \model}: Fig.~(a) illustrates the inference pipeline. \model builds its Connector, Encoder, and Decoder entirely with linear attention in pursuit of efficiency. At each step, the Encoder and Decoder predict the conditioning from the known tokens, after which the denoising network draws sample based on the conditioning.  
    }
    \label{fig:model_inference}
\end{figure}

As described in Fig.~\ref{fig:model_inference}-(a), \model inference pipeline starts by encoding the text prompt with a text encoder and integrating it into the query token through the Connector. 
The image generation process starts with all tokens masked and proceeds through a multi-step, generalized autoregressive procedure that generates image tokens progressively. 
At each step, the Encoder extracts information from the predicted tokens, which—together with the masked tokens—are decoded by the Decoder to form the conditioning. 
A denoising flow matching network (\eg, MLP) samples the token based on this conditioning. 
After all autoregressive steps, the generated tokens are feed into a VAE decoder to produce the final image.

\section{Model Configuration}
\label{sec:appendix.configuration}

The detailed configurations of our \model models of varying sizes for class-conditional image generation are listed in Tab.~\ref{tab:appendix_explore_model_config}, corresponding to the exploration roadmap discussed in Sec.~\ref{sec:scaling}. 
Tab.~\ref{tab:appendix_main_model_config} reports the detailed hyperparameters of \model for both class-conditional and text-to-image generation in our main experiments in Sec.~\ref{sec:experiments}.

\section{Detailed hyper-parameters on ImageNet-1K in Sec.~\ref{sec:scaling}}
\label{sec:appendix.hyper-parameter}

In Tab.~\ref{tab:appendix_scaling_hyperparameter}, we provide the hyper-parameters for the experiments in Sec.~\ref{sec:scaling}, which explore the scaling behavior of different linear attention design choices.

\begin{table}[htbp]
    \centering
    \caption{Detailed \model configurations in Sec.~\ref{sec:scaling}.}
    \label{tab:appendix_explore_model_config}
    \scalebox{0.9}{
    \begin{tabular}{lccc}
    \toprule
    \textbf{Configuration} & \textbf{L} & \textbf{XL} & \textbf{H} \\
    \midrule
    Task & C2I & C2I & C2I  \\
    Resolution & 256px & 256px & 256px  \\
    Params~(M) & 0.4B & 0.6B & 1.4B  \\
    Connector Blocks & 16 & 16 & 16 \\
    Encoder Blocks & 16 & 16 & 16  \\
    Decoder Blocks & 16 & 16 & 16  \\
    Flow Matching MLP Depth & 6 & 6 & 6 \\
    Channels & 768 & 1024 & 1536  \\
    DWC Kernel Size & 5 & 5 & 5  \\
    \bottomrule
    \end{tabular}
    }
\end{table}

\begin{table}[htbp]
    \centering
    \caption{Detailed \model configurations in Sec.~\ref{sec:experiments}.}
    \label{tab:appendix_main_model_config}
    \scalebox{0.9}{
    \begin{tabular}{lcccc}
    \toprule
    \textbf{Configuration} &  \textbf{H} & \textbf{H} & \textbf{H} \\
    \midrule
    Task &  C2I & T2I & T2I \\
    Resolution  & 256px & 512px & 1024px \\
    Params~(M)  & 1.4B & 1.4B & 1.5B \\
    Connector Blocks  & 16 & 16 & 16\\
    Encoder Blocks  & 16 & 16 & 16 \\
    Decoder Blocks  & 16 & 16 & 16 \\
    Flow Matching MLP Depth  & 6 & 6 & 6\\
    Channels  & 1536 & 1536 & 1536 \\
    DWC Kernel Size  & 5 & 5 & 5 \\
    \bottomrule
    \end{tabular}
    }
\end{table}

\begin{table}[htbp]
    \centering
    \caption{Training setting of \model for scaling behavior empirical study in Sec.~\ref{sec:scaling}. }
    \label{tab:appendix_scaling_hyperparameter}
    \scalebox{0.9}{
    \begin{tabular}{lccc}
    \toprule
    \textbf{Training Setting} & \textbf{L} & \textbf{XL} & \textbf{H} \\
    \midrule
    Base Learning Rate & $8\times10^{-4}$ & $8\times10^{-4}$ & $8\times10^{-4}$ \\
    Batch Size & 64$\times$32 & 48$\times$32 & 24$\times$32 \\
    Training Iteration & 200K & 200K & 200K \\
    Weight Decay & 0.02 & 0.02 & 0.02 \\
    Warm-up Steps & 10000 & 10000 & 10000 \\
    Model EMA & 0.99 & 0.99 & 0.99 \\
    \bottomrule
    \end{tabular}
    }
\end{table}

\section{Complexity analysis of DWC Module and KV Gate}
Following VCA~\cite{pu2025linear}, we provide a complexity analysis to the DWC module and KV gate used in \model. 
We follow the notation in Sec.~\ref{sec:preliminary}: assume the model uses $H$ attention heads to process $N=N_{\text{q}}+N_{\text{img}}$ tokens, where each token has a hidden dimension $D$, and the per-head dimension is $d$, satisfying $D = h d$.

\paragraph{Linear attention.}

The computation of the per-token output $O^{(\texttt{d})}_i$ for linear attention with division-based normalization can be expressed as:

\begin{equation}          
\begin{aligned}
O^{(\texttt{d})}_i           
= \frac{\phi(Q_i) \left(\sum_{j=1}^{N}{\phi(K_j)^\top V_j }\right)}{\phi(Q_i) \left(\sum_{m=1}^{N}{\phi(K_m)^\top}\right)},
\label{eq:re:dbla}
\end{aligned}
\end{equation}

Theoretically, for the numerator, we have:
\begin{equation}          
\begin{aligned}
\underbrace{\phi(Q_i)}_{\mathbb{R}^{1\times d}} \left(\sum_{j=1}^{N}{\underbrace{\phi(K_j)^\top}_{\mathbb{R}^{d\times 1}} \underbrace{V_j}_{\mathbb{R}^{1\times d}} }\right)
\;\longrightarrow\;
 \mathbb{R}^{1\times d},
\label{eq:re:dbla-num}
\end{aligned}
\end{equation}

where, first, a $d \times 1$ matrix is multiplied by a $1 \times d$ matrix, which costs $\mathcal{O}(d^{2})$.  
Then, a $1 \times d$ matrix is multiplied by an a $d \times d$ matrix, which also costs $\mathcal{O}(d^{2})$.

For the denominator, we have:
\begin{equation}          
\begin{aligned}
\underbrace{\phi(Q_i)}_{\mathbb{R}^{1\times d}} \left(\sum_{m=1}^{N}{\underbrace{\phi(K_m)^\top}_{\mathbb{R}^{d\times 1}}}\right)
\;\longrightarrow\;
 \mathbb{R}^{1\times 1},
\label{eq:re:dbla-denom}
\end{aligned}
\end{equation}

where, a $1 \times d$ matrix is multiplied by a $d \times 1$ matrix, which costs $\mathcal{O}(d)$, which is negligible in the big-$\mathcal{O}$ sense.

Computing the final output $O^{(\texttt{d})}_i$ also costs $\mathcal{O}(d)$, which is negligible in the big-$\mathcal{O}$ sense.  
Thus, the per-token complexity is at most $\mathcal{C}_{la, t}=2d^{2}=\mathcal{O}(d^{2})$.

For a single attention head processing $N$ tokens, the total cost is at most:
         
\begin{align}
\text{Linear attention, per-head:}\qquad\mathcal{C}_{la, h}=N\mathcal{C}_{la, t}= \mathcal{O}(N d^{2}).
\label{eq:re:cost:head}
\end{align}

For the whole linear attention with $h$ heads (and hidden dimension $D = h d$), the total cost is:
\begin{align}
\text{Linear attention:}\qquad\mathcal{C}_{la}=h\mathcal{C}_{la, h}= \mathcal{O}(NDd).
\label{eq:re:cost:attention}
\end{align}

\paragraph{Depthwise convolution.}

The DWC module applies a $k \times k$ depthwise convolution to $N_{\text{img}}$ image tokens only, with a cost of:

\begin{align}
\text{DWC module:}\qquad\mathcal{C}_{dwc}= \mathcal{O}(N_{\text{img}}Dk^2).
\label{eq:re:cost:dwc}
\end{align}

The ratio between the computational cost of the \textit{DWC module} and that of \textit{linear attention} is:

\begin{equation}
\frac{\mathcal{O}(N_{\text{img}}Dk^2)}{\mathcal{O}(NDd)}\leq\frac{k^2}{d}.
\end{equation}
Note that the DWC module processes only $N_{\text{img}}$ image tokens, while the linear attention processes both $N_{\text{q}}$ query tokens and $N_{\text{img}}$ image tokens ($N=N_{\text{q}}+N_{\text{img}}$).

For our text-to-image model LINA-H, we have $D = 1536$ and $h = 16$, so the per-head dimension is $d = 96$. Therefore, we obtain $k^{2}/d = 0.26$.

Note that the actual cost ratio is \textbf{smaller} than $0.26$.
The reasons are as follows. 
In \model, the DWC module is applied only in the \textit{decoder} blocks, which account for merely $\frac{1}{3}$ of the total network depth. As a result, the additional computational overhead introduced by DWC is non-dominant.

% is used only in the \textit{Decoder} blocks, and these Decoder blocks constitute \textit{only $\frac{1}{3}$ of the total number of blocks}. Therefore, the DWC module introduces only non-dominant additional computational cost.

\paragraph{KV gate.}

The KV gate $g^{(k)},g^{(v)}\in\mathbb{R}^{N_{\text{img}}}$ uses learnable parameters to scale the keys and values on a per-token basis.

\begin{equation}          
\begin{aligned}
\tilde{K}_j=\underbrace{g_j^{(k)}}_{\mathbb{R}^{1\times 1}}\underbrace{\phi(K_j)}_{\mathbb{R}^{1\times d}}, ~~
\tilde{V}_j=\underbrace{g_j^{(v)}}_{\mathbb{R}^{1\times 1}}\underbrace{V_j}_{\mathbb{R}^{1\times d}},
\label{eq:re:cost:kvgate}
\end{aligned}
\end{equation} 

For a single head processing $N_{\text{img}}$ tokens, the cost is at most:

\begin{align}
\text{KV gate, per-head:}\qquad\mathcal{C}_{kvg,h}= 2N_{\text{img}}d= \mathcal{O}(N_{\text{img}}d).
\label{eq:re:cost:kvgate:head}
\end{align}

For the whole linear attention with $h$ heads (and hidden dimension $D = h d$), the total cost is:
\begin{align}
\text{KV gate:}\qquad\mathcal{C}_{kvg}= h\mathcal{C}_{kvg,h}=\mathcal{O}(N_{\text{img}}D).
\label{eq:re:cost:kvgate}
\end{align}

The ratio between the computational cost of the \textit{KV gate} and that of \textit{linear attention} is:

\begin{equation}
\frac{\mathcal{O}(N_{\text{img}}D)}{\mathcal{O}(NDd)}\leq\frac{1}{d}.
\end{equation}
Note that the KV gate processes only $N_{\text{img}}$ image tokens, while the linear attention processes both $N_{\text{q}}$ query tokens and $N_{\text{img}}$ image tokens ($N=N_{\text{q}}+N_{\text{img}}$).

Since the per-head dimension is $d = 96$, the additional computational cost introduced by the KV gate is negligible.

\paragraph{Comparison to softmax attention.}
The computational complexity of standard softmax attention is $\mathcal{C}_{fa}=O(N^2D)$, and the ratio between linear attention and softmax attention is:
\begin{align}
\frac{\mathcal{C}_{la}}{\mathcal{C}_{fa}}
= \frac{\mathcal{O}(N D d)}{\mathcal{O}(N^2D)}=\frac{d}{N}.
\label{eq:re:cost:ratio}
\end{align}

Note that when LINA operates at 1024px, we have $N = 5120$, $D = 1536$, and $h = 16$, so $d = D/h = 96$. This gives a complexity ratio of $d/N = 96/5120 \approx 0.019$. Therefore, linear attention achieves a substantial reduction in computational cost, which is consistent with our measured FLOPs.

\section{How Our Findings Relate to Prior Work on Linear Attention}

We note that neither the normalization types used in linear attention nor the locality augmentation techniques are our original inventions. However, to the best of our knowledge, our work is the \textit{first} to systematically investigate these components in the context of \textit{autoregressive image generation}. Prior studies such as Flattened Transformer~\cite{han2023flatten} and InLine~\cite{han2024bridging} only explored them in perception tasks such as image classification.

To clarify the relationship between our findings and existing results in other domains, we summarize the key similarities and differences below.

\paragraph{Differences from visual perception tasks.}

In autoregressive image generation, our results show that \textit{division-based normalization} yields better \textit{scaling behavior} for linear attention than subtraction-based normalization. This observation is not fully aligned with the conclusions drawn from perception tasks (\eg, InLine). We suspect that this discrepancy may arise from the inherent gap between generation and perception tasks. Issues such as ``semantic confusion'', highlighted by InLine, may not be the main bottleneck in generative models. Instead, division-based normalization might be inherently more suitable for denoising-based generative processes. We admit that we do not yet have a theoretical explanation for this phenomenon. Nonetheless, we believe that the empirical results and the new data points we provide will be valuable for guiding future work.

\paragraph{Similarities to visual perception tasks.}

Consistent with findings in perception tasks (\eg, Flattened Transformer), our autoregressive generation results confirm that \textit{linear attention indeed benefits from additional locality modeling}. This can be seen in the scaling behavior in Fig.~\ref{fig:scaling_behavior}-(a) in the main paper.
Nevertheless, our work further raises two conceptual questions in Sec.~\ref{sec:scaling}, \ie,
1) Whether the softmax operation is the key reason behind the locality gap between softmax attention and linear attention; and
2) Whether locality modeling universally helps autoregressive image generation.
Although our conclusions overlap with prior work in image classification, we believe these discussions offer useful conceptual insights into the role of locality in linear attention for autoregressive generation.

In summary, our results suggest that \textit{linear attention in autoregressive image generation comes with task-specific considerations}, such as preferring division-based normalization, rather than directly inheriting conclusions from perception tasks. We hope our study can provide reliable guidelines for future research, helping to avoid large amounts of repetitive ablations.

\section{Scaling Behavior: Detailed Results}
\label{sec:appendix.scaling-results}

In Tab.~\ref{tab:appendix_scaling_result}, we present detailed results, as discussed in Sec.~\ref{sec:scaling}, comparing the scaling behavior of the four linear attention design choices. 
For evaluation metrics, we literally reported include FID-50K~\citep{heusel2017gans}, sFID~\citep{nash2021generating}, Inception Score~\citep{salimans2016improved}, and Precision/Recall~\citep{kynkaanniemi2019improved}. 

\begin{table}[htbp]
    \centering
    \caption{Scaling performance of different linear attention design choices. We report the results on the class-conditional image generation using the ImageNet benchmark at 256$\times$256 resolution.}
    \label{tab:appendix_scaling_result}
    \scalebox{0.88}{
    \begin{tabular}{lccccc}
    \toprule
    \textbf{Linear Attention Setting} & \textbf{FID$\downarrow$ (w/o cfg)} & \textbf{sFID$\downarrow$}& \textbf{IS$\uparrow$} & \textbf{Precision$\uparrow$} & \textbf{Recall$\uparrow$}\\
    \midrule
    \textit{Large (L)}:  & & & & & \\
    Division-based Normalization, w/o DWC & 13.13  & 7.58  & 105.26  & 0.65  & 0.61  \\
    Subtraction-based Normalization, w/o DWC & 15.81  & 8.95  & 90.47  & 0.62  & 0.60  \\
    Division-based Normalization, w/ DWC & 9.11  & 5.89  & 117.40  & 0.69  & 0.61  \\
    Subtraction-based Normalization, w/ DWC & 9.02  & 5.77  & 116.54  & 0.70  & 0.61  \\
    \midrule
    \textit{Extra Large (XL)}:  & & & & & \\
    Division-based Normalization, w/o DWC & 9.73  & 6.46  & 121.68  & 0.68  & 0.60  \\
    Subtraction-based Normalization, w/o DWC & 12.10  & 7.50  & 107.07  & 0.66  & 0.60  \\
    Division-based Normalization, w/ DWC & 7.25  & 5.17  & 127.93  & 0.71  & 0.61  \\
    Subtraction-based Normalization, w/ DWC & 7.48  & 5.38  & 125.80  & 0.72  & 0.60  \\
    \midrule
    \textit{Huge (H)}:  & & & & & \\
    Division-based Normalization, w/o DWC & 9.87  & 6.05  & 115.64  & 0.69  & 0.58  \\
    Subtraction-based Normalization, w/o DWC & 11.10  & 7.08  & 107.25  & 0.67  & 0.59  \\
    Division-based Normalization, w/ DWC & 6.53  & 5.01  & 130.31  & 0.72  & 0.60  \\
    Subtraction-based Normalization, w/ DWC & 7.88  & 5.28  & 117.26  & 0.72  & 0.58  \\
    \bottomrule
    \end{tabular}
    }
\end{table}

\begin{table}[htbp]
    \centering
    \caption{Training setting of our text-to-image generation \model. }
    \label{tab:appendix_t2i_hyperparameter}
    \scalebox{0.97}{
    \begin{tabular}{lccc}
    \toprule
    \textbf{Training Setting} & \textbf{Stage 1} & \textbf{Stage 2} & \textbf{Stage 3} \\
    \midrule
    Training Iterations & 565K & 600K & 50K \\
    Dataset Size & 28M & 28M & 16M \\
    Resolution & 256px & 512px & 1024px \\
    Base Learning Rate & $2e^{-4}$ & $1e^{-4}$ & $5e^{-5}$ \\
    Batch Size & 16$\times$48 & 4$\times$48 & 1$\times$48 \\
    Weight Decay & 0.02 & 0.02 & 0.02 \\
    Warm-up Steps & 10000 & 10000 & 10000 \\
    Model EMA & 0.99 & 0.99 & 0.99 \\
    \bottomrule
    \end{tabular}
    }
\end{table}

\section{Detailed hyper-parameters on Text-to-image Generation}
\label{sec:appendix.t2isetting}

In Tab.~\ref{tab:appendix_t2i_hyperparameter}, we present the training setting for the text-to-image generation experiments in Sec.~\ref{sec:experiments}, including the dataset size, batch size, learning rate, \etc. The training process can be divided into three stages. Our training is conducted using 48 NVIDIA A100 (40G) GPUs.

\section{KV Gate}
\label{sec:appendix.kvgate-results}

\subsection{Ablation of KV gate Designs}
The variants of the KV gate ablation are listed below.

\paragraph{Mode 1 (\textit{KV gate}).}
\begin{equation}
\small
\begin{gathered}
{\tilde{K}_j}={g_j^{(k)}}\phi(K_j), ~~
{\tilde{V}_j}={g_j^{(v)}}V_j,~~\textit{for}~~j\in \left[1,N\right]
\\
M=\sum_{j=1}^{N}{ 
{\tilde{K}_j}^\top{\tilde{V}_j} 
}=
\sum_{j=1}^{N}{{g_j^{(k)}}{g_j^{(v)}}M_j},~~
z=\sum_{m=1}^{N}{{\tilde{K}_m}^\top},~~
O^{(\texttt{d})}_i          
= \frac{\phi(Q_i) M}{\phi(Q_i) z}.
\label{eq:kvgate_mode1}
\end{gathered}
\end{equation} 
where ${g^{(k)}}, {g^{(v)}}\in\mathbb{R}^N$ are learnable parameters.

\paragraph{Mode 2 (\textit{K gate}).}
\begin{equation}
\small
\begin{gathered}
{\tilde{K}_j}={g_j^{(k)}}\phi(K_j), ~~\textit{for}~~j\in \left[1,N\right]
\\
M=\sum_{j=1}^{N}{ 
{\tilde{K}_j}^\top V_j 
}=
\sum_{j=1}^{N}{{g_j^{(k)}}M_j},~~
z=\sum_{m=1}^{N}{{\tilde{K}_m}^\top},~~
O^{(\texttt{d})}_i          
= \frac{\phi(Q_i) M}{\phi(Q_i) z}.
\label{eq:kvgate_mode2}
\end{gathered}
\end{equation} 
where ${g^{(k)}}\in\mathbb{R}^N$ are learnable parameters.

\paragraph{Mode 3 (\textit{V gate}).}
\begin{equation} 
\small
\begin{gathered}
{\tilde{V}_j}={g_j^{(v)}}V_j,~~\textit{for}~~j\in \left[1,N\right]
\\
M=\sum_{j=1}^{N}{ 
\phi(K_j)^\top{\tilde{V}_j} 
}=
\sum_{j=1}^{N}{{g_j^{(v)}}M_j},~~
z=\sum_{m=1}^{N}{\phi(K_m)^\top},~~
O^{(\texttt{d})}_i          
= \frac{\phi(Q_i) M}{\phi(Q_i) z}.
\label{eq:kvgate_mode3}
\end{gathered}
\end{equation} 
where ${g^{(v)}}\in\mathbb{R}^N$ are learnable parameters.

\paragraph{Mode 4 (\textit{KV gate} + extra $z$ gate).}
\begin{equation} 
\small
\begin{gathered}
{\tilde{K}_j}={g_j^{(k)}}\phi(K_j), ~~{\bar{K}_j}={g_j^{(n)}}\phi(K_j), ~~
{\tilde{V}_j}={g_j^{(v)}}V_j,~~\textit{for}~~j\in \left[1,N\right]
\\
M=\sum_{j=1}^{N}{ 
{\tilde{K}_j}^\top{\tilde{V}_j} 
}=
\sum_{j=1}^{N}{{g_j^{(k)}}{g_j^{(v)}}M_j},~~
z=\sum_{m=1}^{N}{{\bar{K}_m}^\top},~~
O^{(\texttt{d})}_i          
= \frac{\phi(Q_i) M}{\phi(Q_i) z}.
\label{eq:kvgate_mode4}
\end{gathered}
\end{equation} 
where ${g^{(k)}}, {g^{(v)}}, {g^{(n)}}\in\mathbb{R}^N$ are learnable parameters. 

\subsection{Visualization}

\begin{figure*}[t]
    \centering

    \begin{subfigure}{0.23\textwidth}
        \includegraphics[width=\linewidth]{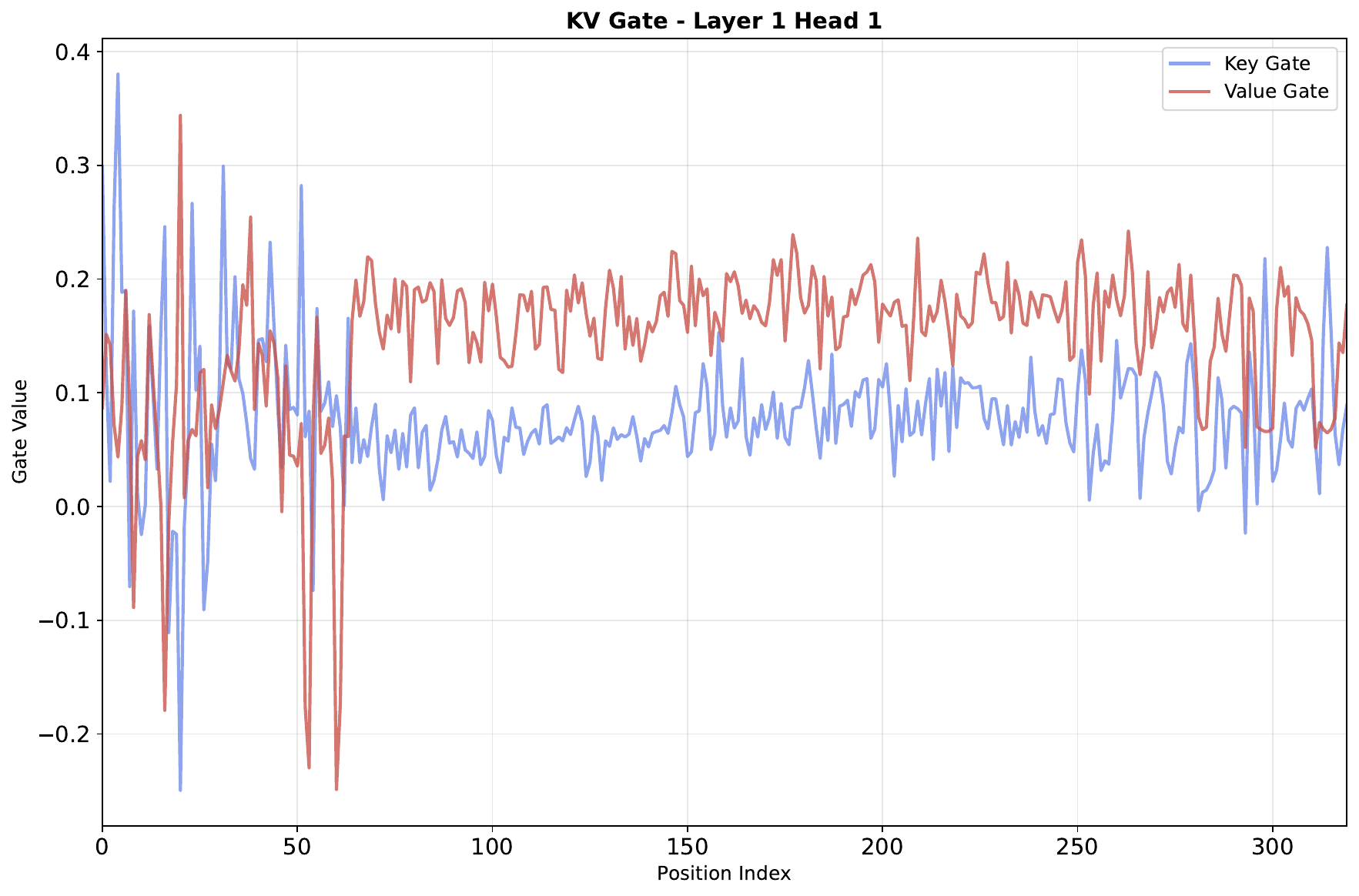}
        \caption{layer 1, head 1}
    \end{subfigure}
    \hspace{0.02\textwidth}
    \begin{subfigure}{0.23\textwidth}
        \includegraphics[width=\linewidth]{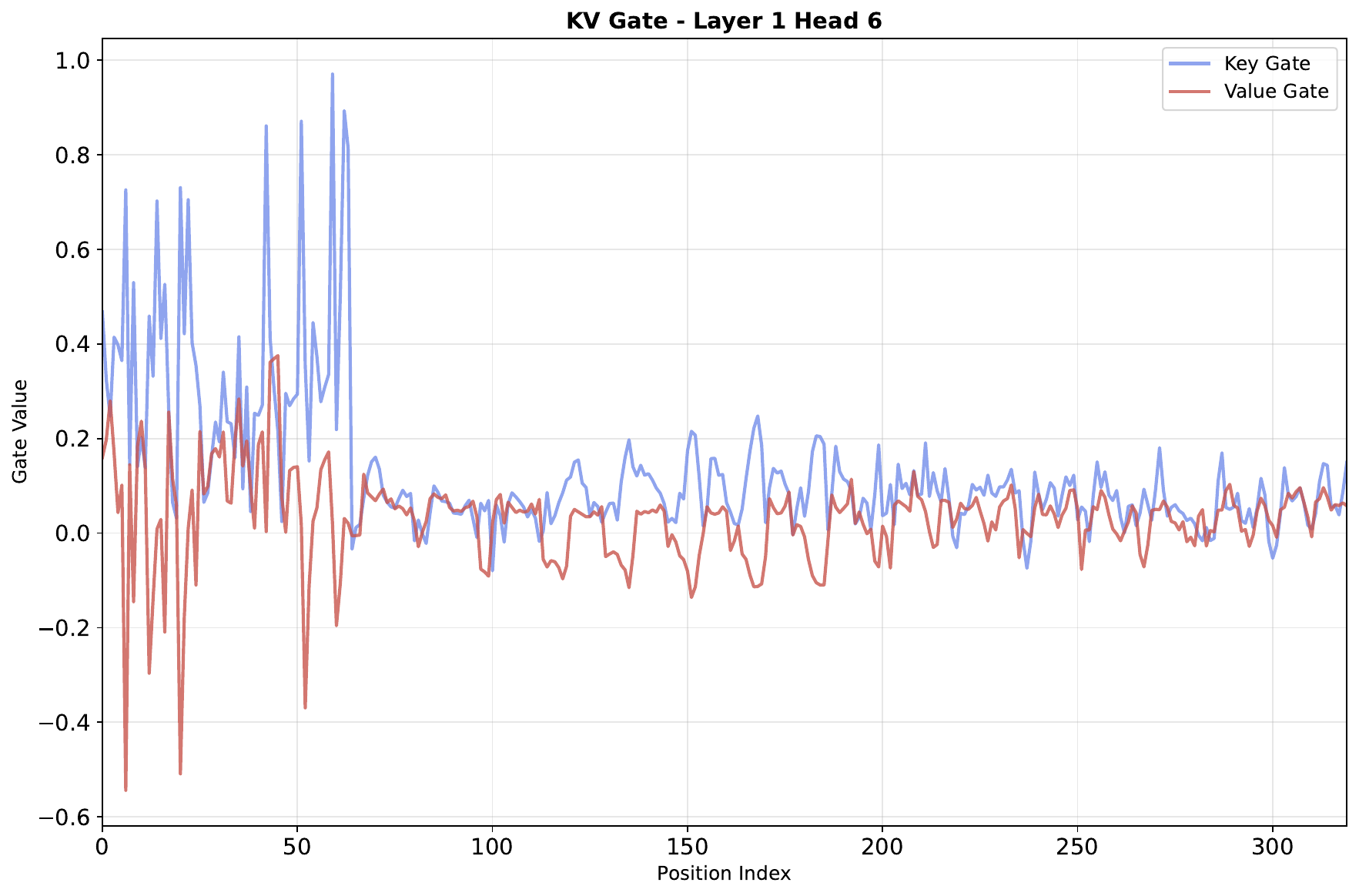}
        \caption{layer 1, head 6}
    \end{subfigure}
    \hspace{0.02\textwidth}
    \begin{subfigure}{0.23\textwidth}
        \includegraphics[width=\linewidth]{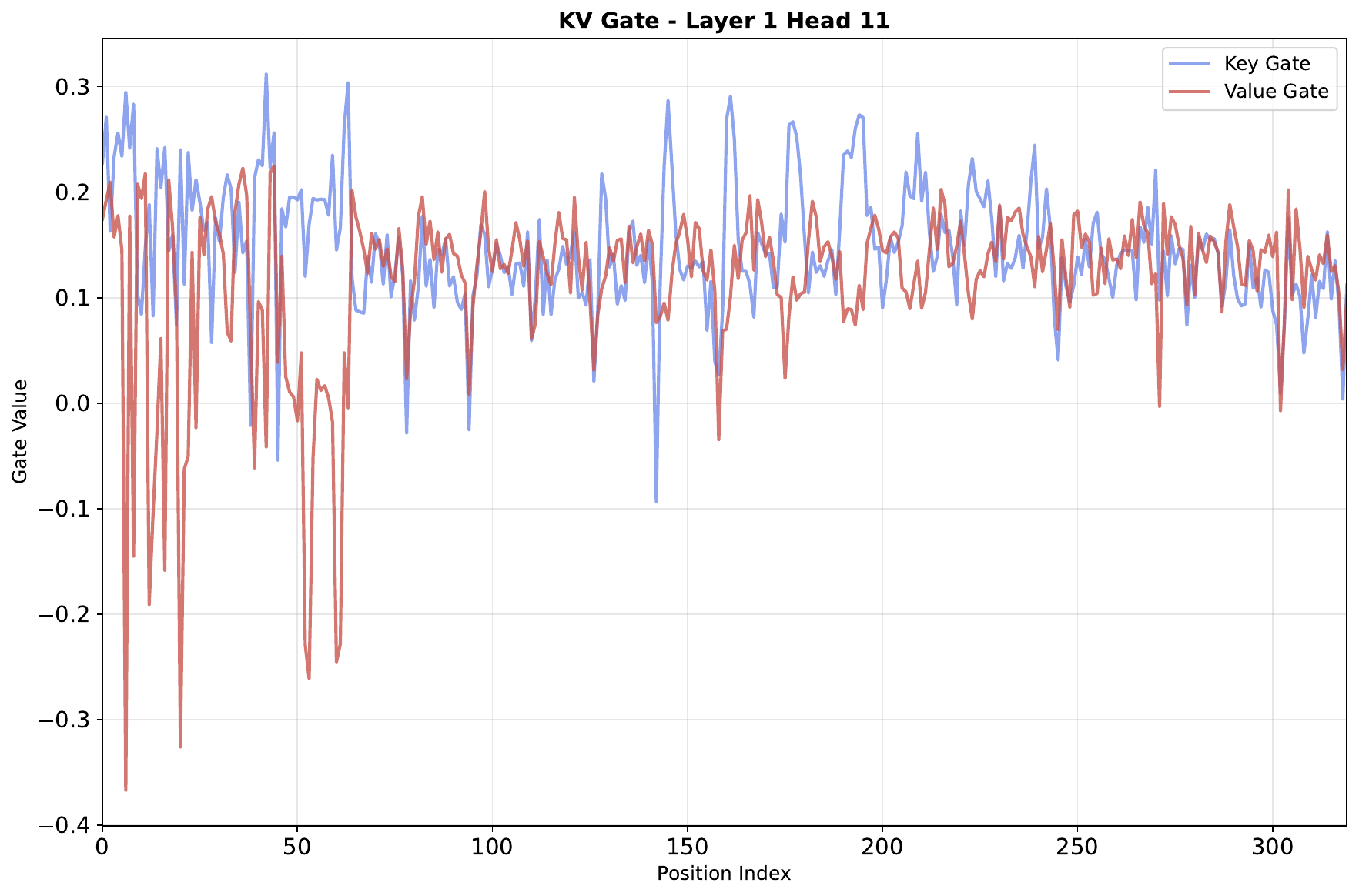}
        \caption{layer 1, head 11}
    \end{subfigure}

    \vspace{0.15cm}

    \begin{subfigure}{0.23\textwidth}
        \includegraphics[width=\linewidth]{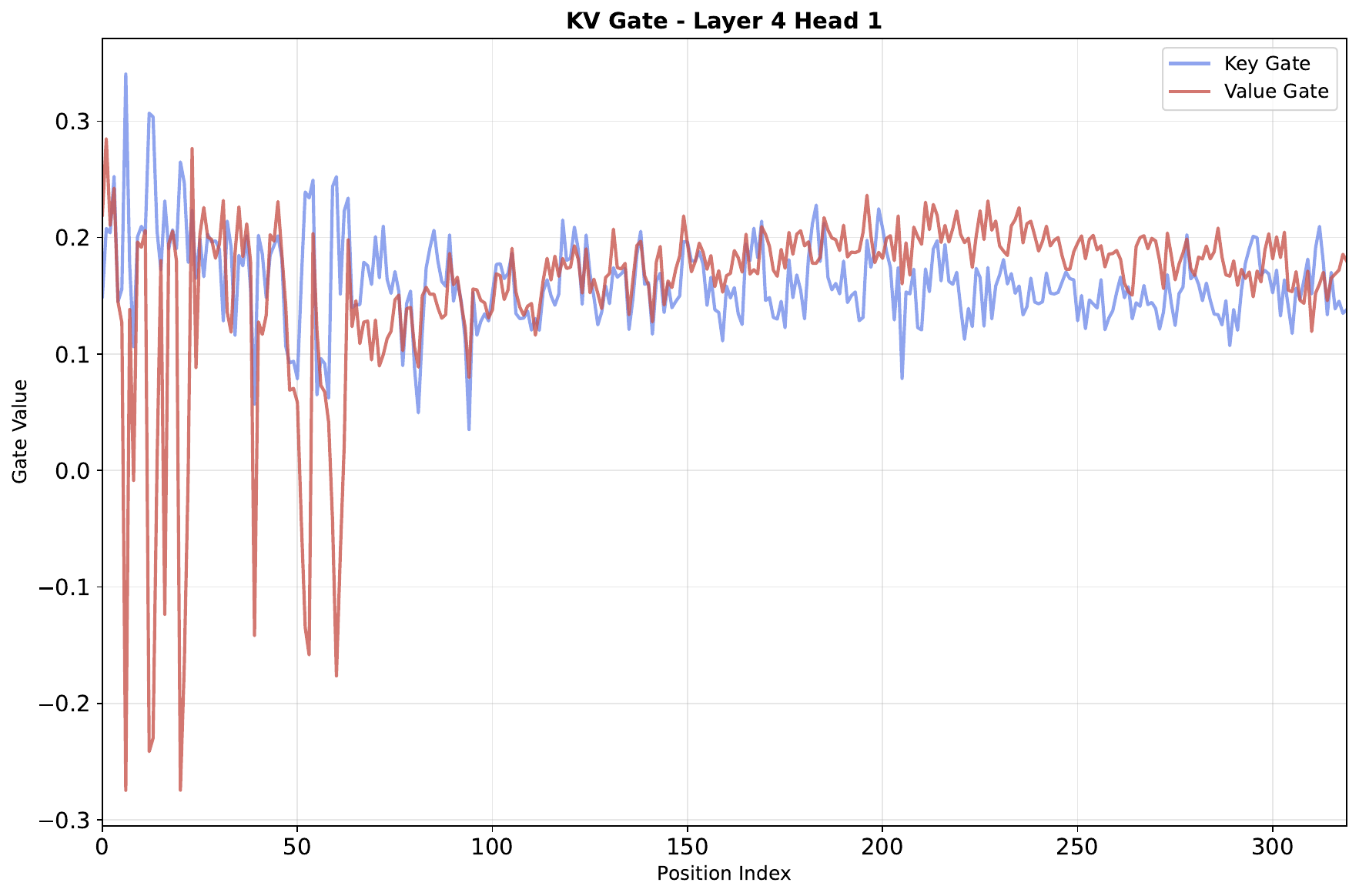}
        \caption{layer 4, head 1}
    \end{subfigure}
    \hspace{0.02\textwidth}
    \begin{subfigure}{0.23\textwidth}
        \includegraphics[width=\linewidth]{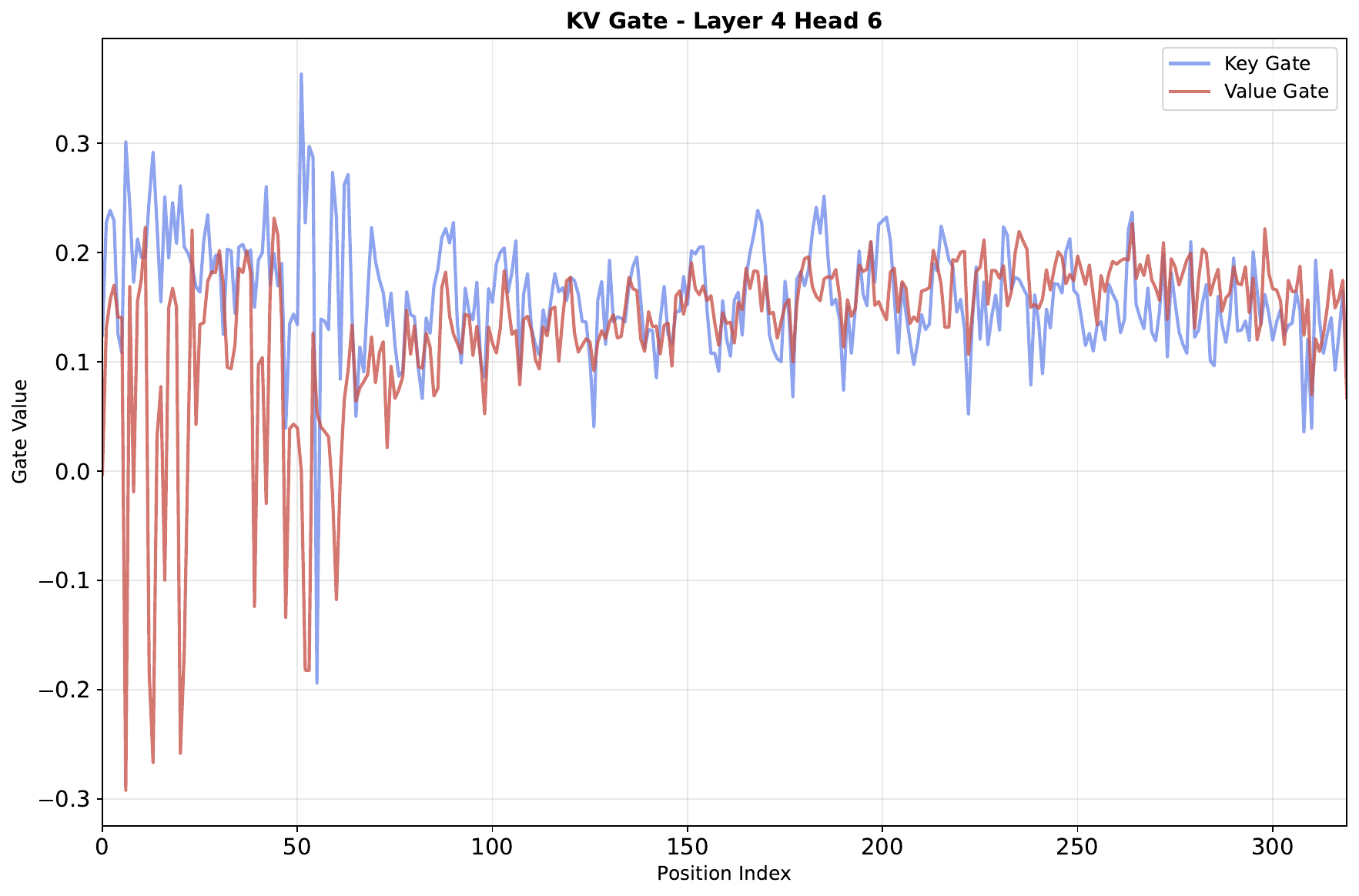}
        \caption{layer 4, head 6}
    \end{subfigure}
    \hspace{0.02\textwidth}
    \begin{subfigure}{0.23\textwidth}
        \includegraphics[width=\linewidth]{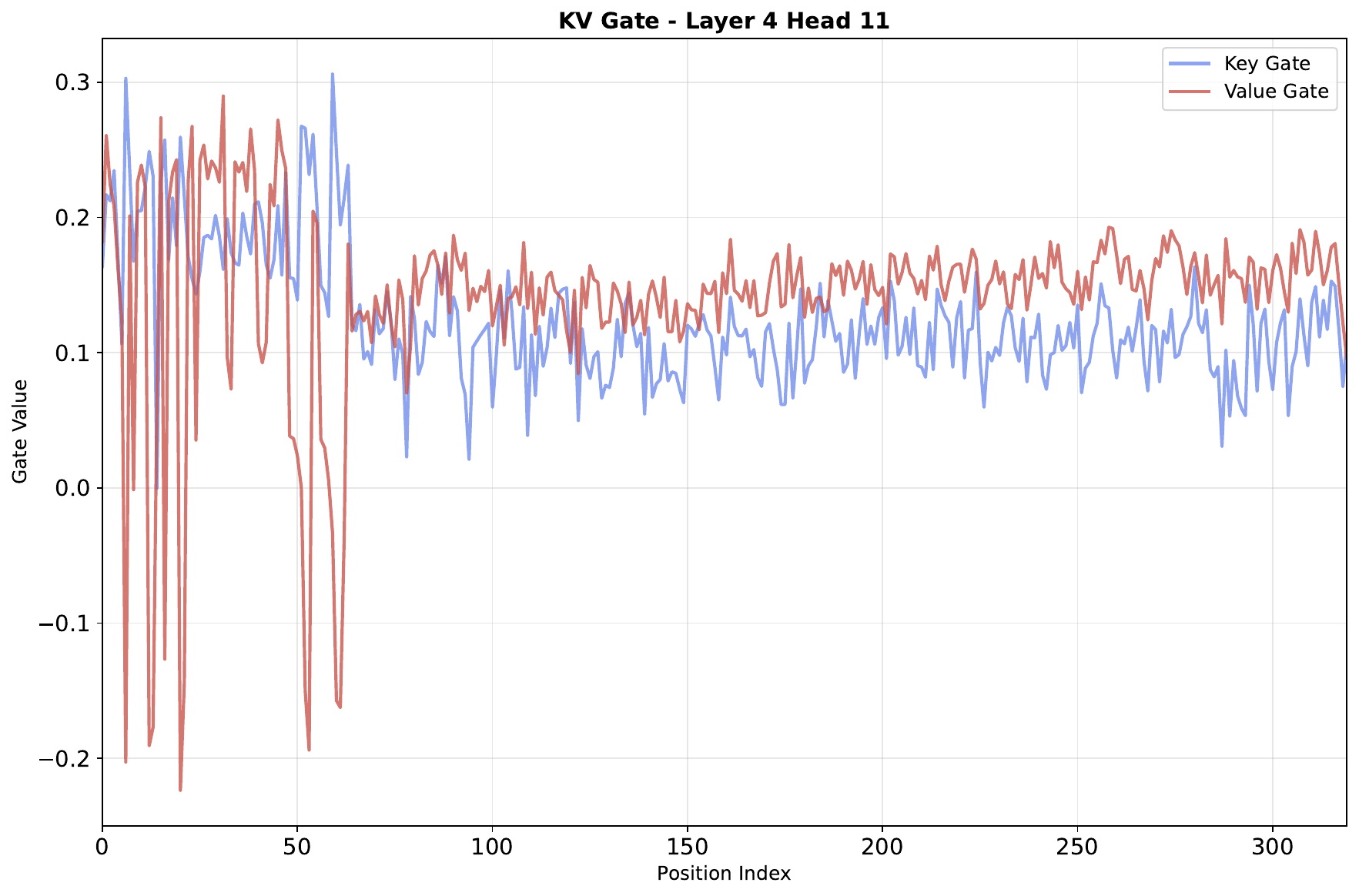}
        \caption{layer 4, head 11}
    \end{subfigure}

    \vspace{0.15cm}

    \begin{subfigure}{0.23\textwidth}
        \includegraphics[width=\linewidth]{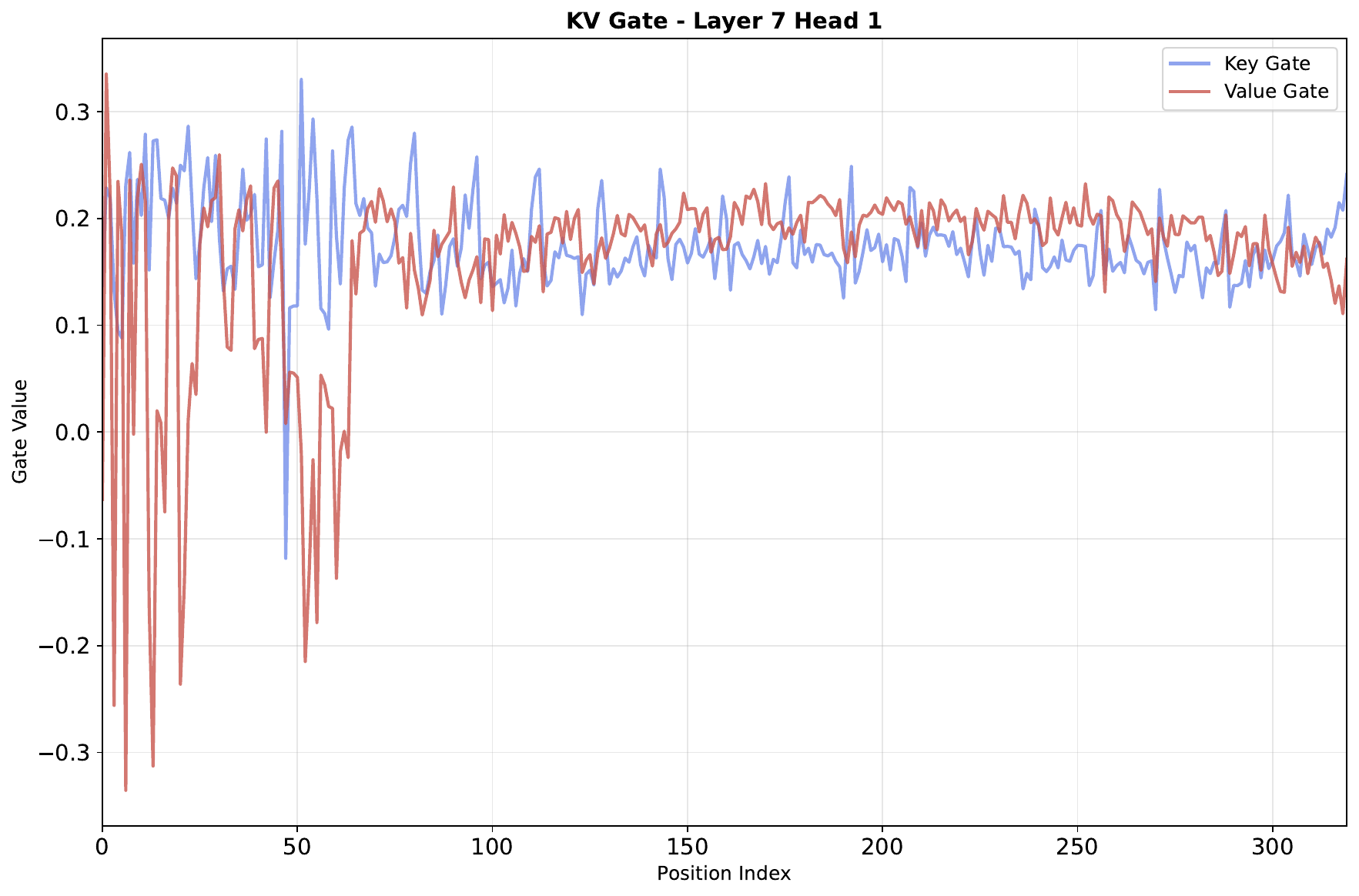}
        \caption{layer 7, head 1}
    \end{subfigure}
    \hspace{0.02\textwidth}
    \begin{subfigure}{0.23\textwidth}
        \includegraphics[width=\linewidth]{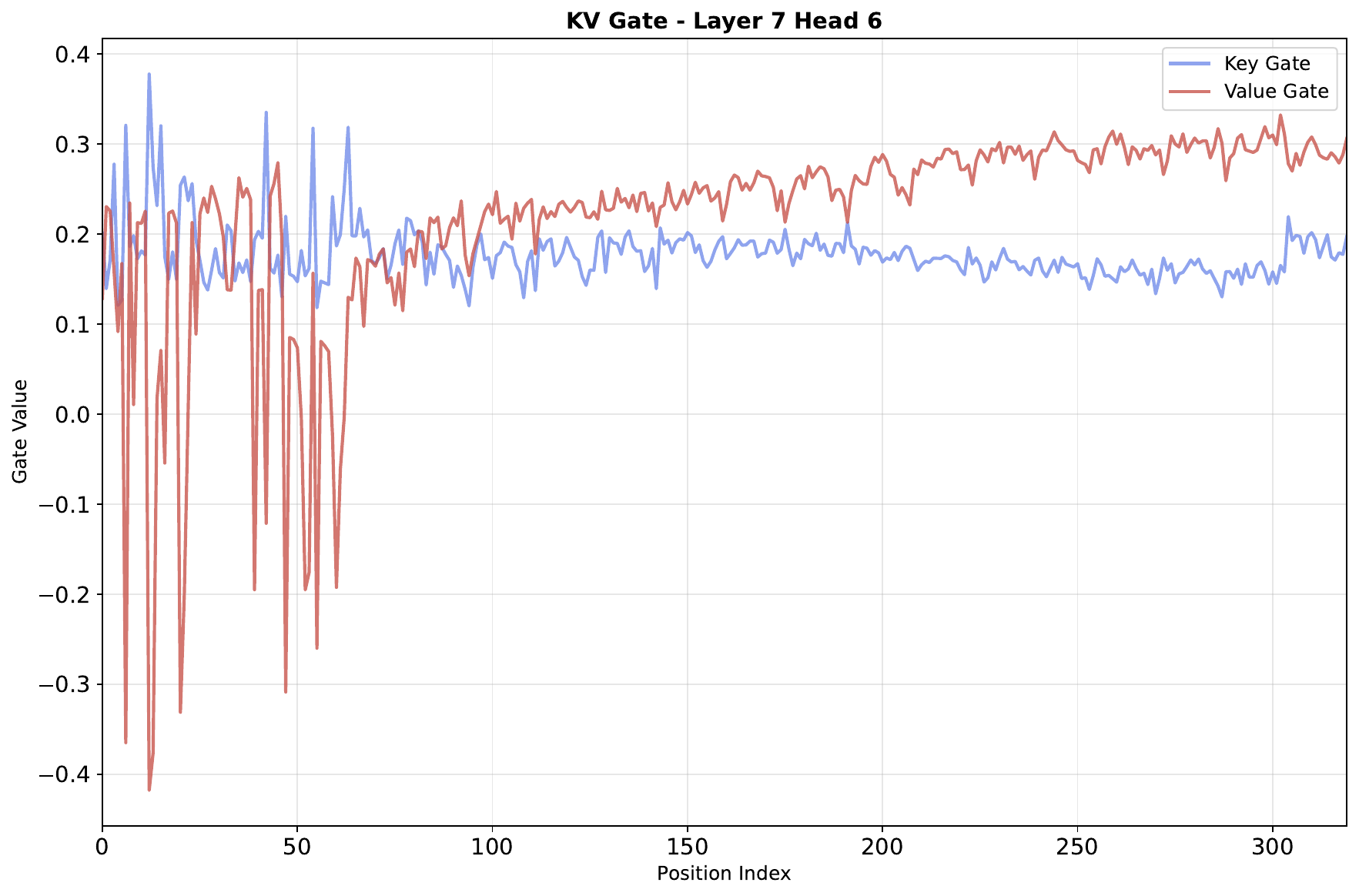}
        \caption{layer 7, head 6}
    \end{subfigure}
    \hspace{0.02\textwidth}
    \begin{subfigure}{0.23\textwidth}
        \includegraphics[width=\linewidth]{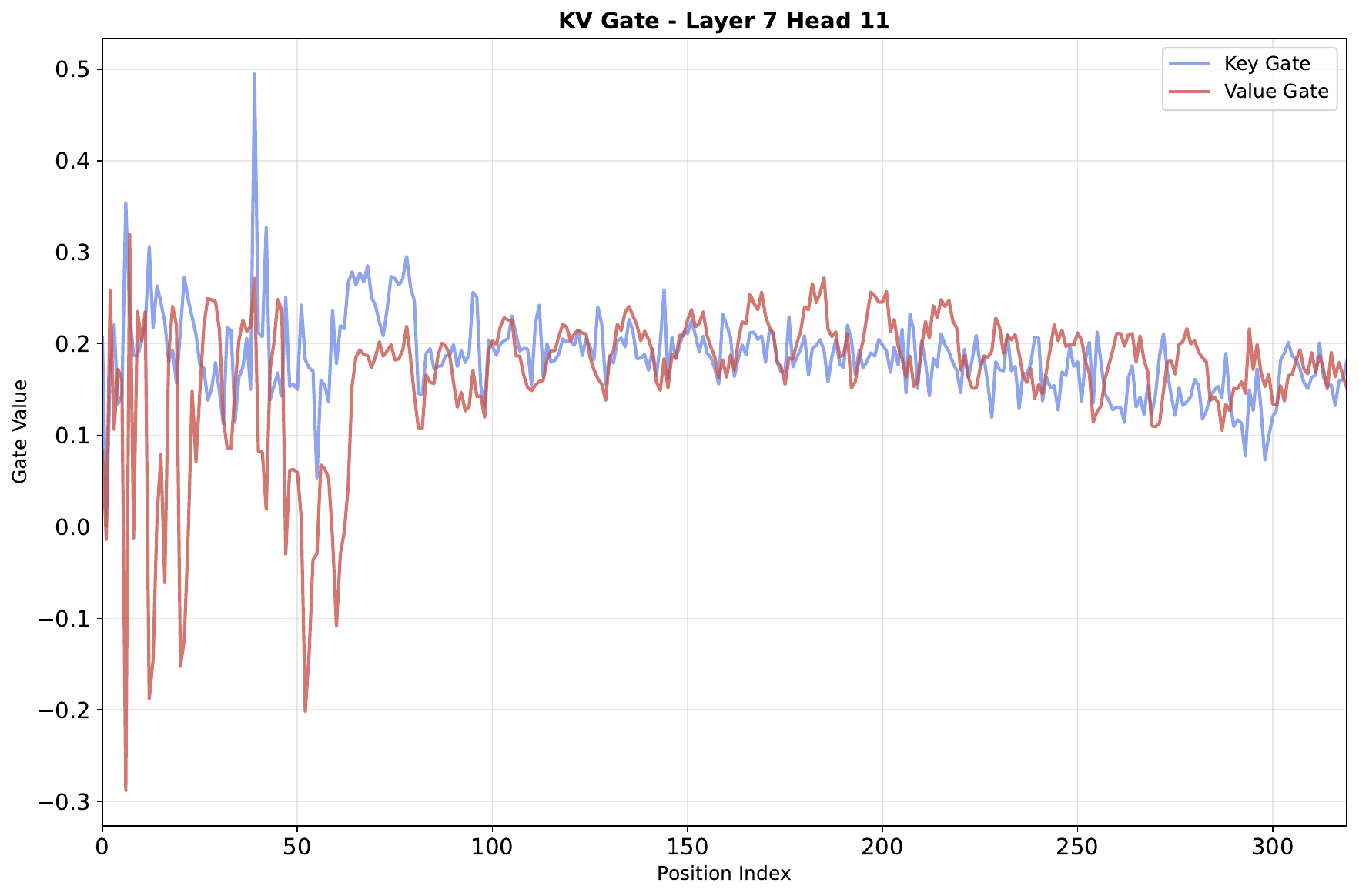}
        \caption{layer 7, head 11}
    \end{subfigure}

    \vspace{0.15cm}

    \begin{subfigure}{0.23\textwidth}
        \includegraphics[width=\linewidth]{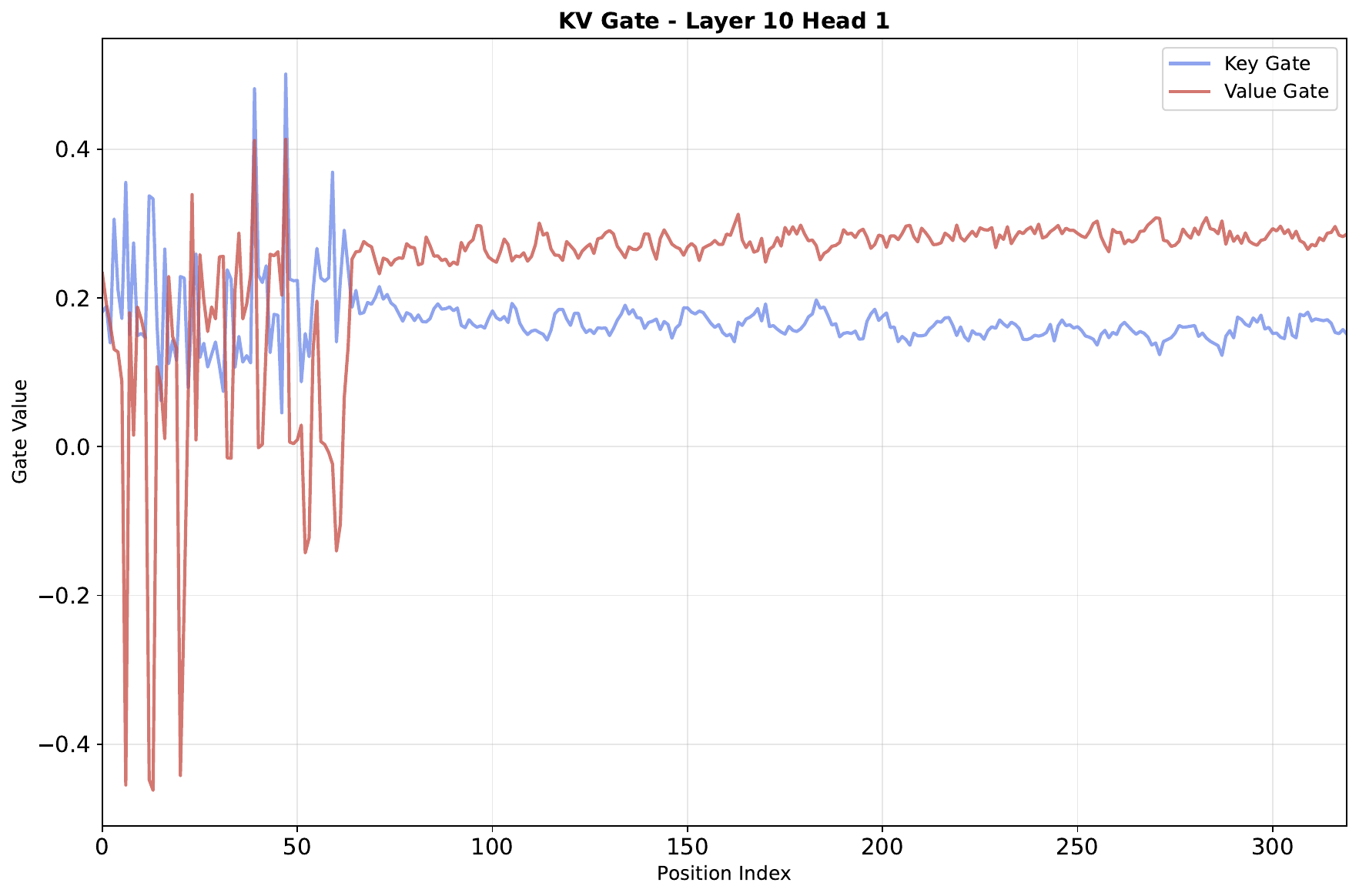}
        \caption{layer 10, head 1}
    \end{subfigure}
    \hspace{0.02\textwidth}
    \begin{subfigure}{0.23\textwidth}
        \includegraphics[width=\linewidth]{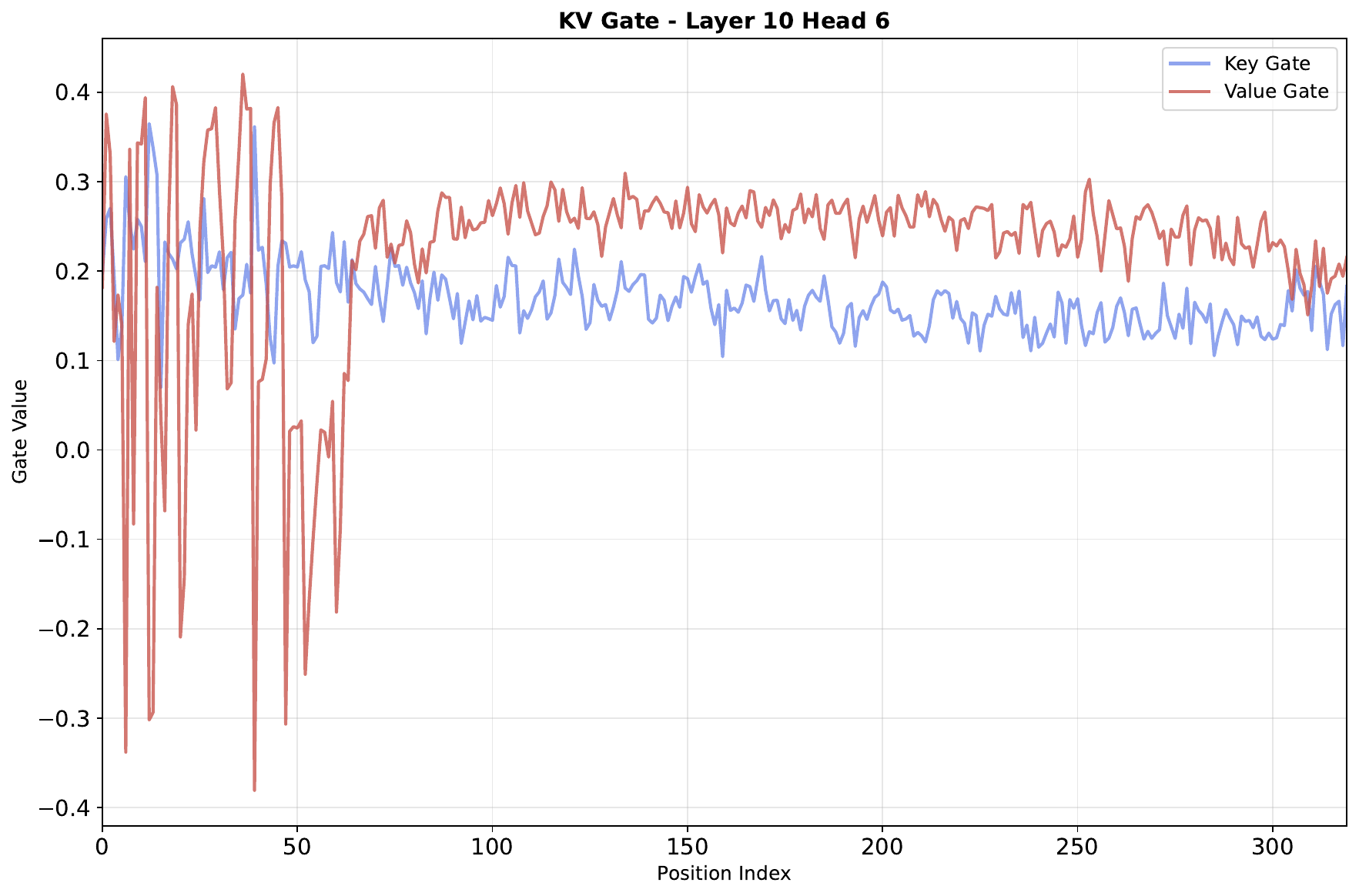}
        \caption{layer 10, head 6}
    \end{subfigure}
    \hspace{0.02\textwidth}
    \begin{subfigure}{0.23\textwidth}
        \includegraphics[width=\linewidth]{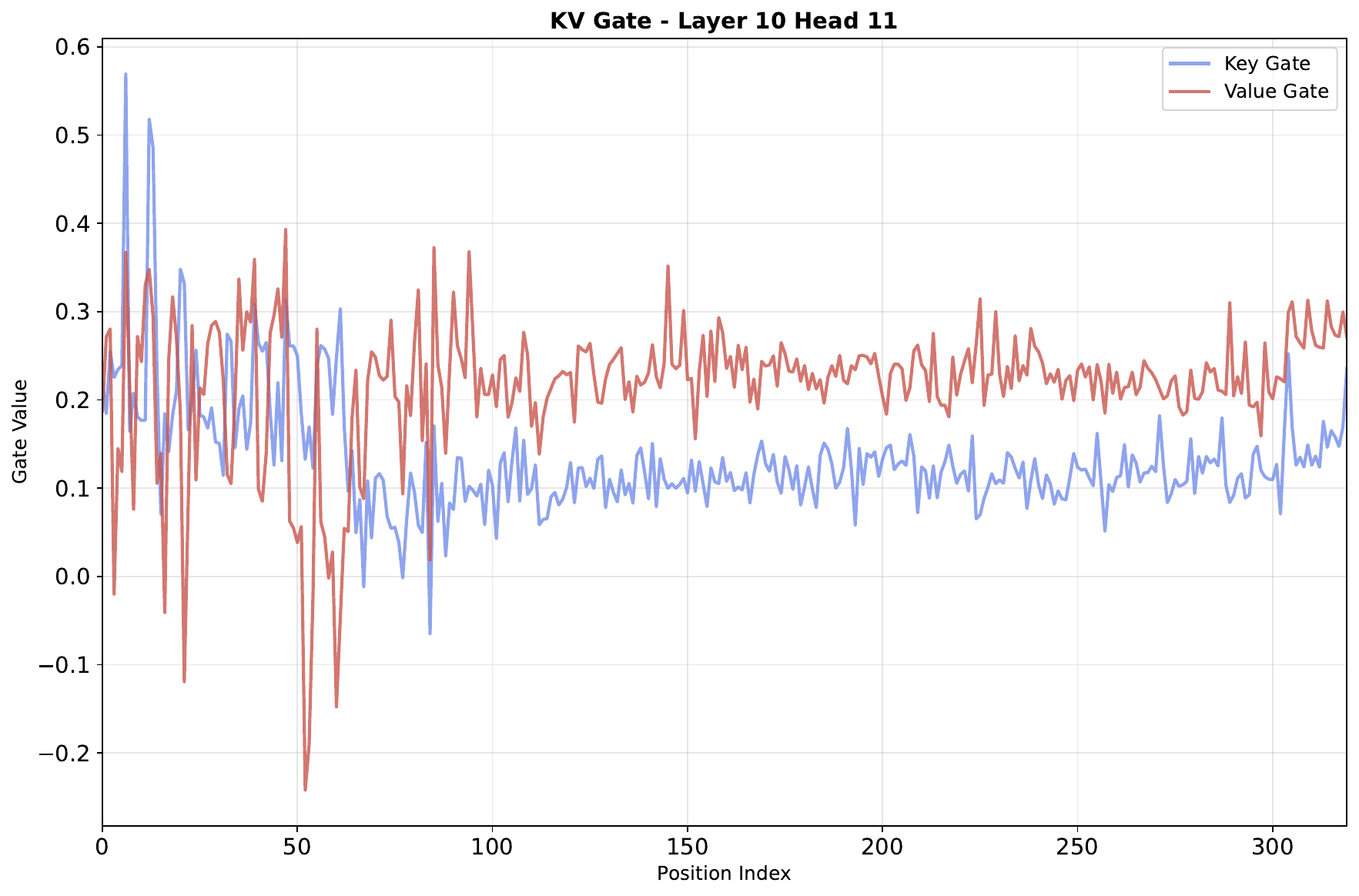}
        \caption{layer 10, head 11}
    \end{subfigure}

    \vspace{0.15cm}

    \begin{subfigure}{0.23\textwidth}
        \includegraphics[width=\linewidth]{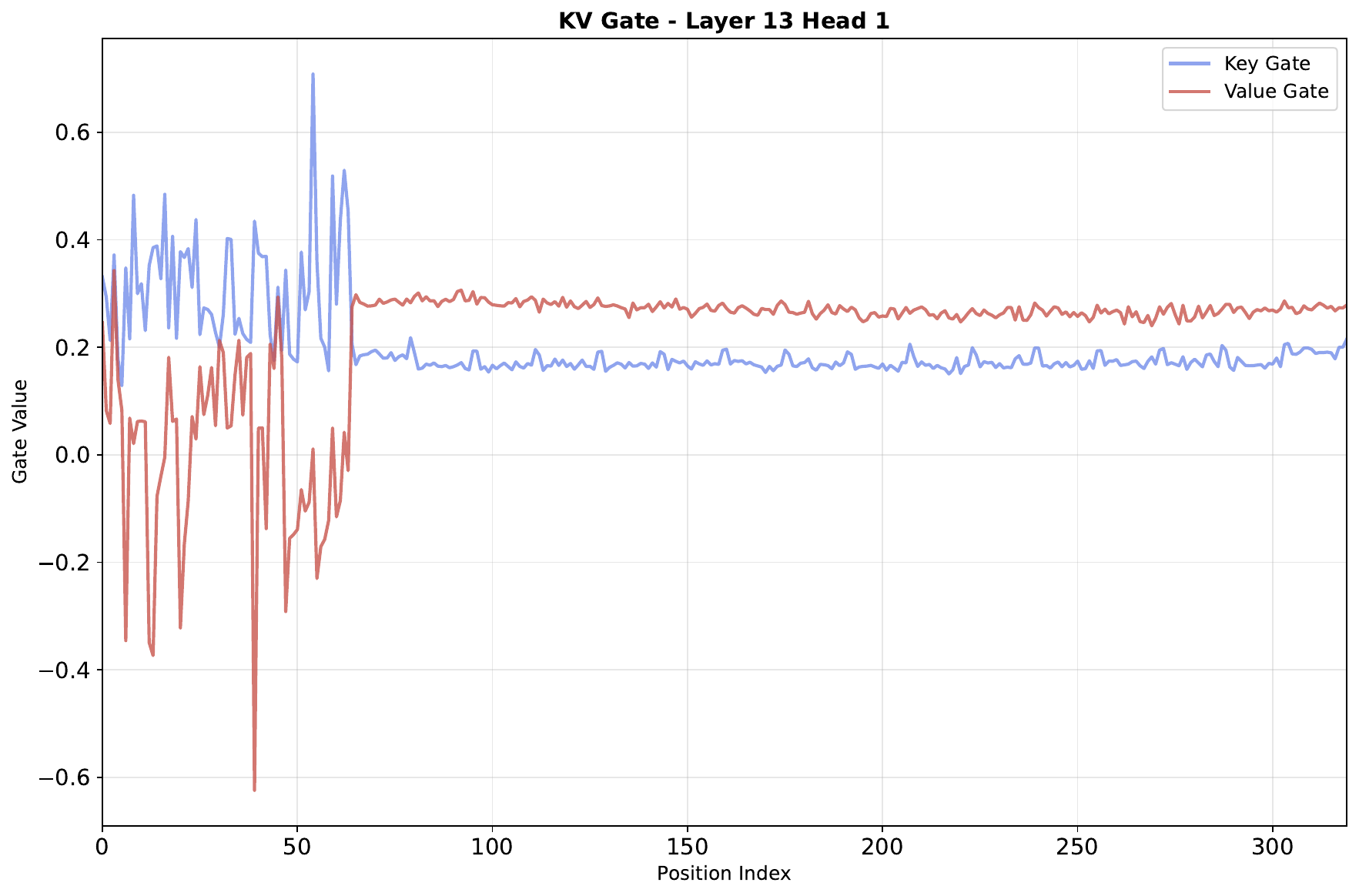}
        \caption{layer 13, head 1}
    \end{subfigure}
    \hspace{0.02\textwidth}
    \begin{subfigure}{0.23\textwidth}
        \includegraphics[width=\linewidth]{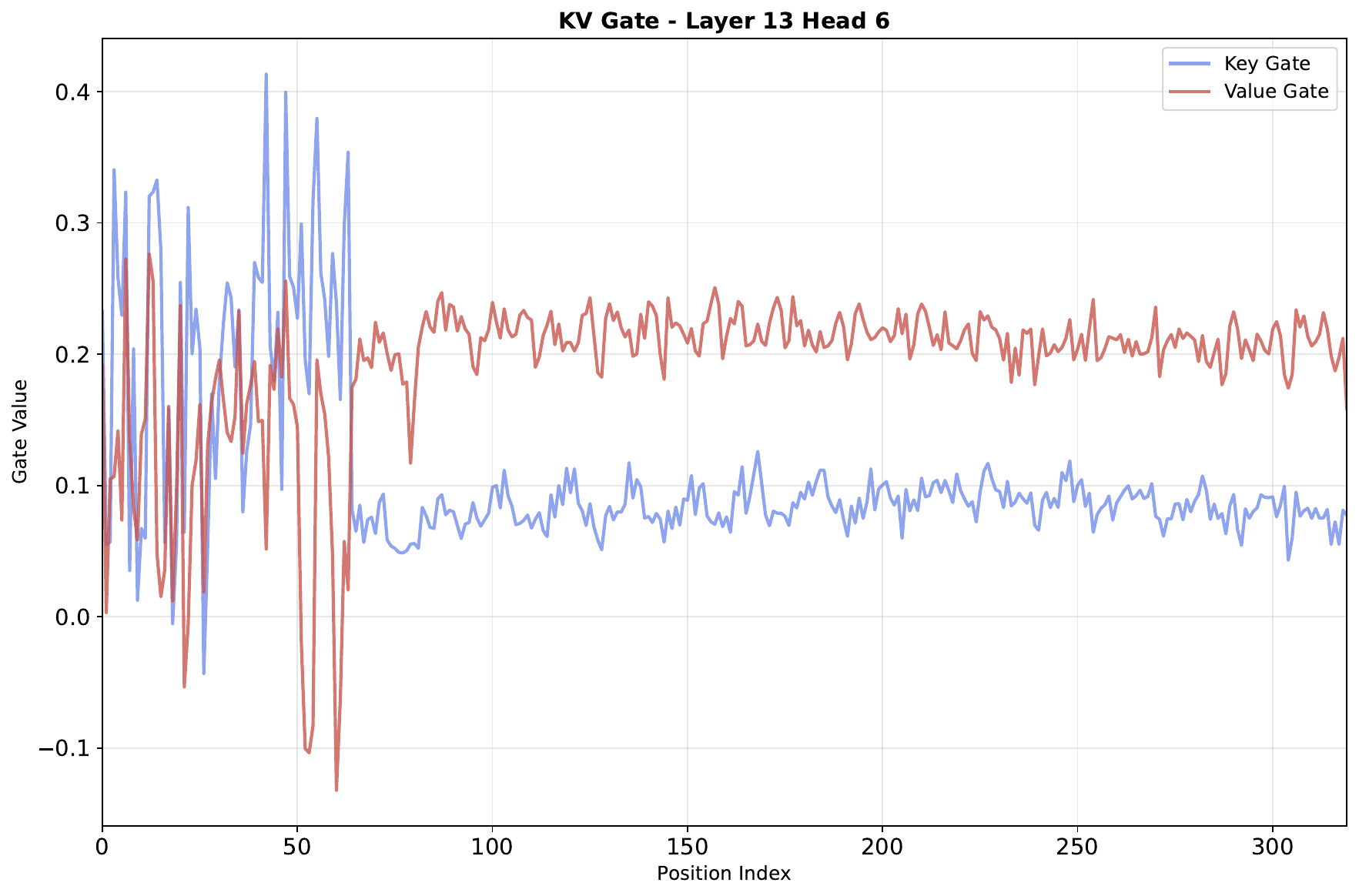}
        \caption{layer 13, head 6}
    \end{subfigure}
    \hspace{0.02\textwidth}
    \begin{subfigure}{0.23\textwidth}
        \includegraphics[width=\linewidth]{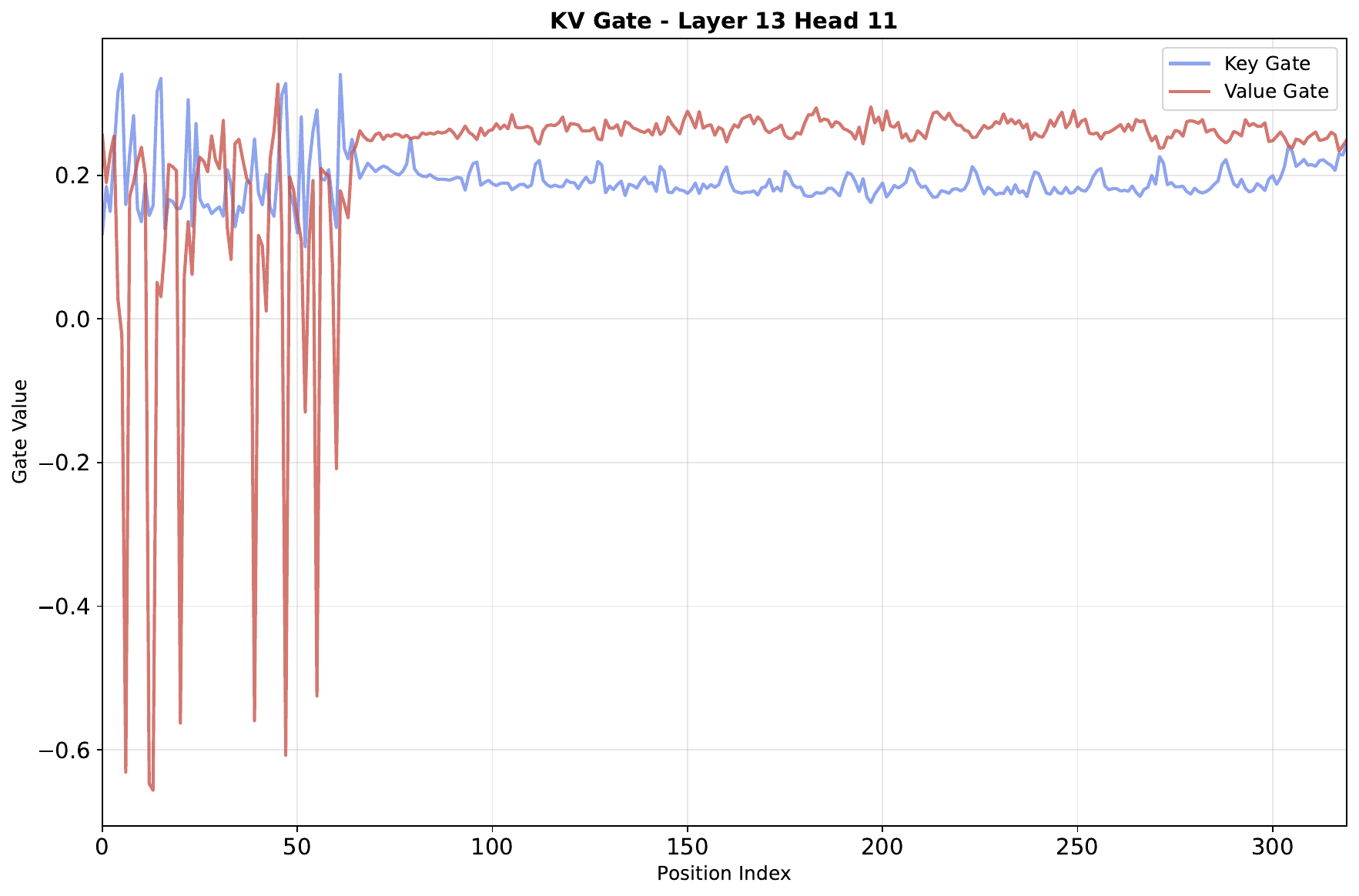}
        \caption{layer 13, head 11}
    \end{subfigure}
    
    \vspace{0.15cm}

    \begin{subfigure}{0.23\textwidth}
        \includegraphics[width=\linewidth]{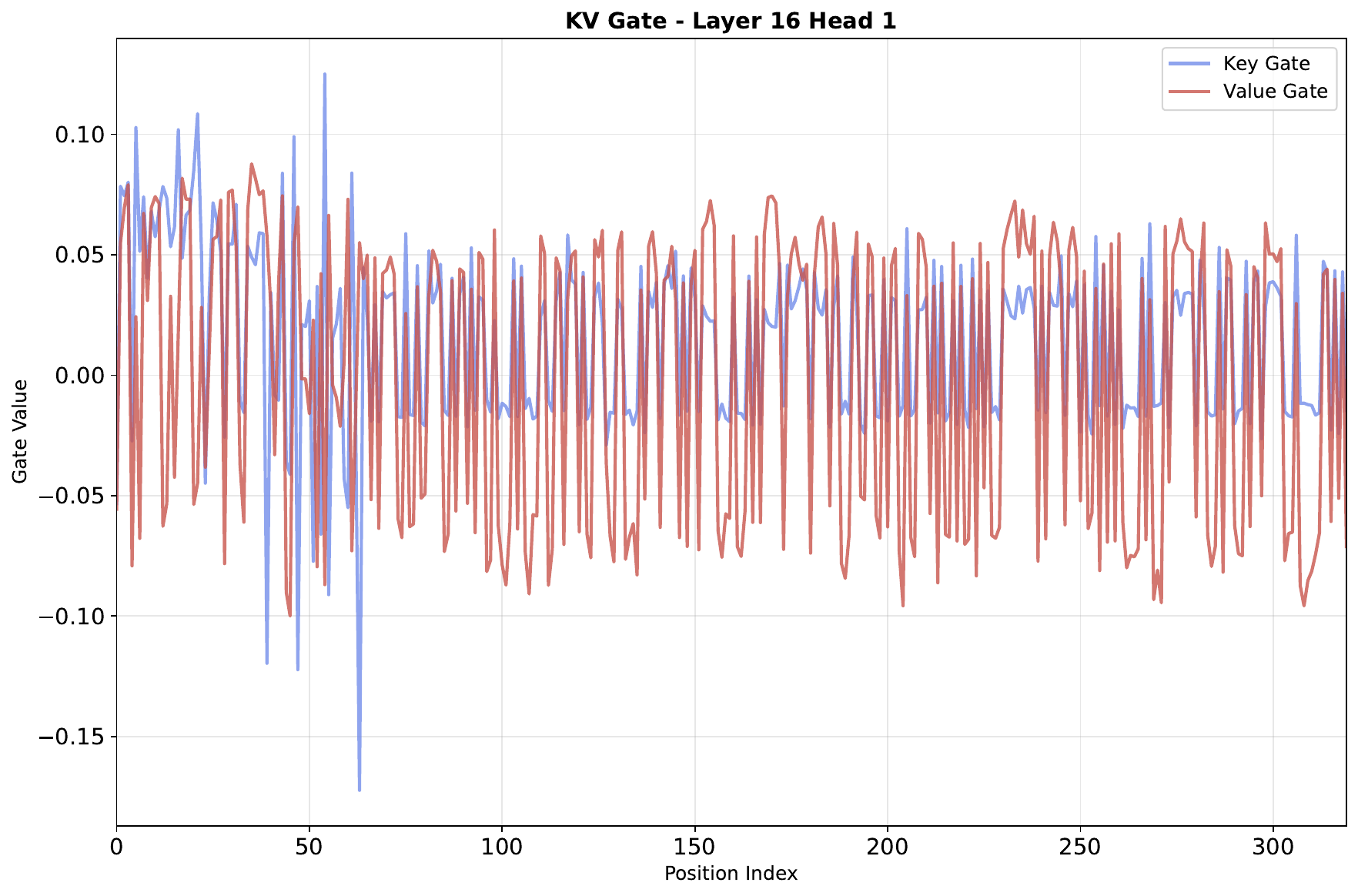}
        \caption{layer 16, head 1}
    \end{subfigure}
    \hspace{0.02\textwidth}
    \begin{subfigure}{0.23\textwidth}
        \includegraphics[width=\linewidth]{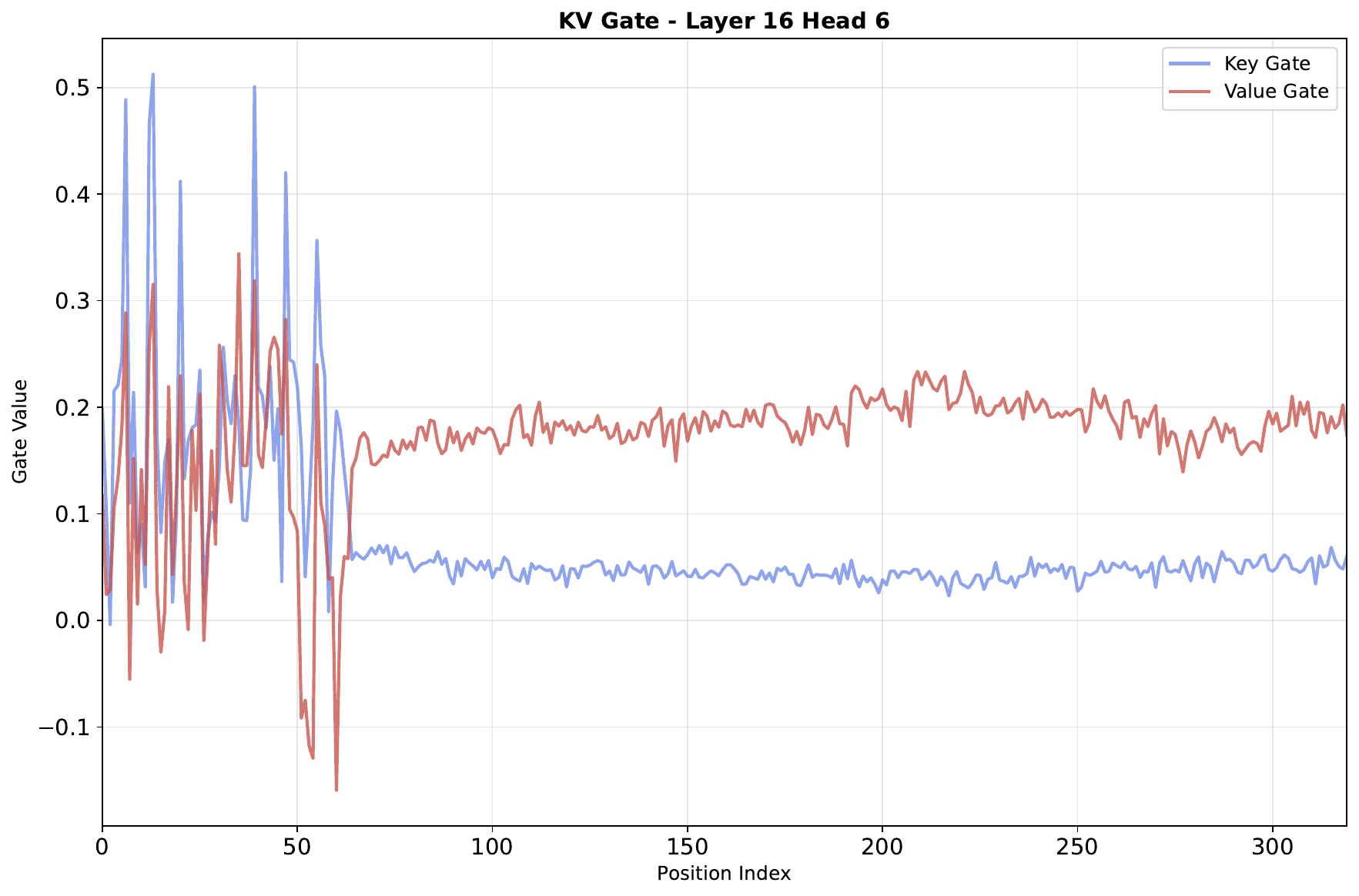}
        \caption{layer 16, head 6}
    \end{subfigure}
    \hspace{0.02\textwidth}
    \begin{subfigure}{0.23\textwidth}
        \includegraphics[width=\linewidth]{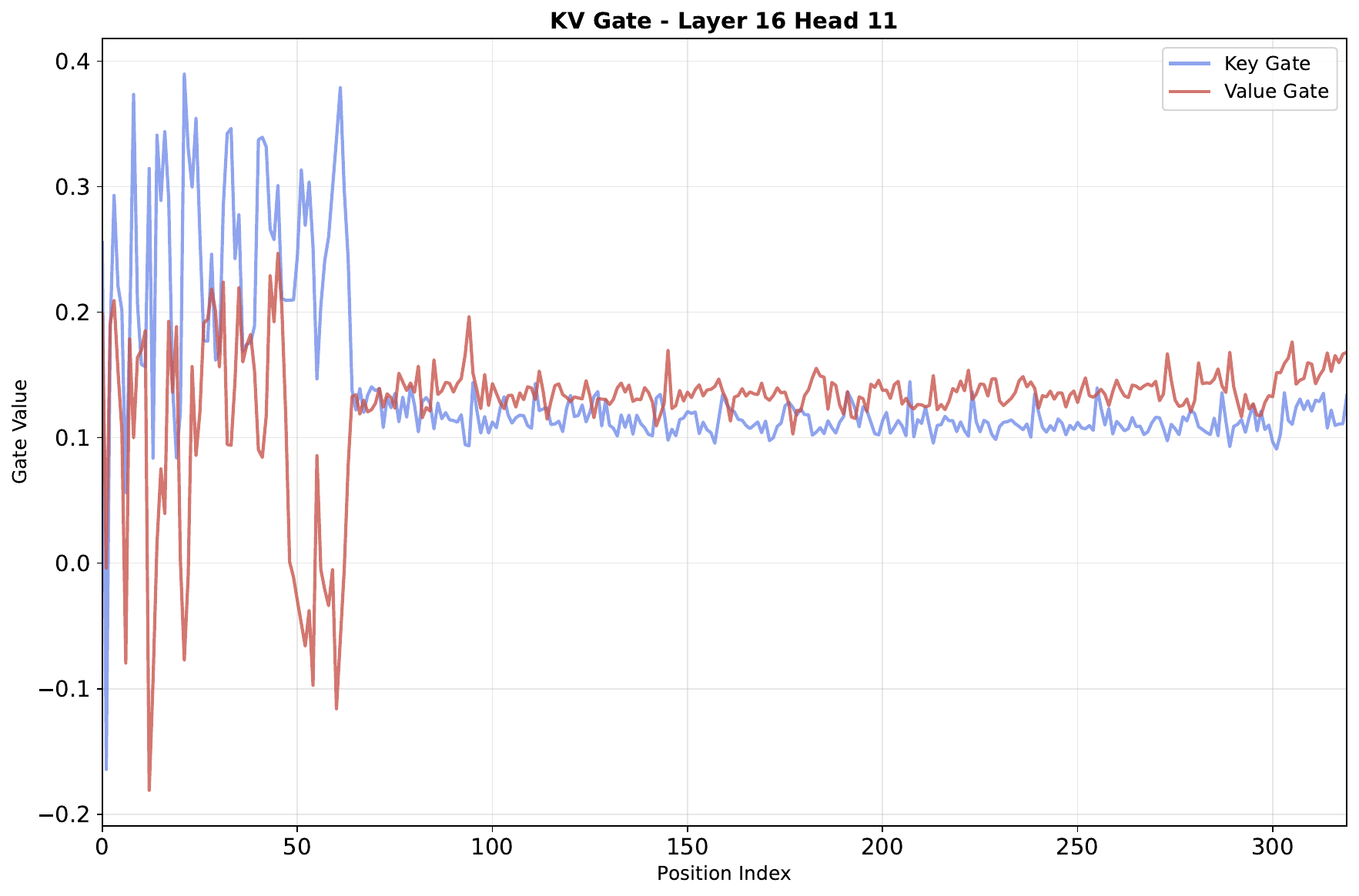}
        \caption{layer 16, head 11}
    \end{subfigure}

    \caption{\textbf{\textit{KV gate} visualization.} We plot the \textit{KV gate} results for layers 1, 4, 7, 10, 13, 16 (indexed 1–16) and heads 1, 6, 11 (indexed 0–15). Across different layers and heads, the \textit{KV gate} learns distinct patterns, allowing flexible memory management. Zoom in for best view.} 
\label{fig:kvgate-visualion}
\end{figure*}

Figure~\ref{fig:kvgate-visualion} presents detailed visualizations of the learned KV gate values across different layers and heads. 
From the visualizations, we find a common pattern across layers and heads: the first 64 query tokens show fluctuations, while the following 256 image tokens remain relatively stable with distinct behaviors. We leave the analysis of text tokens to future work and focus here on the KV gate patterns for image tokens. 

Our observations are three-folds: 
(1) From a cross-layer perspective, the KV gate patterns for image tokens also differ by layer. For example, in layer 1, head 1 the fluctuations are more pronounced, indicating substantial variation in KV gate values across image tokens. In contrast, in layer 13, head 1 the fluctuations are much weaker, suggesting that the values remain relatively stable across tokens. 
(2) From the perspective of value ranges, in many layers the KV gate values lie mostly within $(-1, 1)$, such as layer 10, head 6, layer 4, head 1, and layer 4, head 11. This suggests that, in certain cases, the model applies token-wise attenuation to both $M$ and $z$ when regulating memory in linear attention. 
(3) From the perspective of heads, the KV gate within the same layer seems to exhibit diverse patterns across different heads. For example, in layer 16, the fluctuations of head 6 and head 11 seems relatively stable, whereas head 1 seems to oscillate more strongly.

\section{More Qualitative Results}
\label{sec:appendix.qualitative}
\begin{figure}[htbp]
    \vspace{0em}
    \centering
    \includegraphics[width=0.99\textwidth]{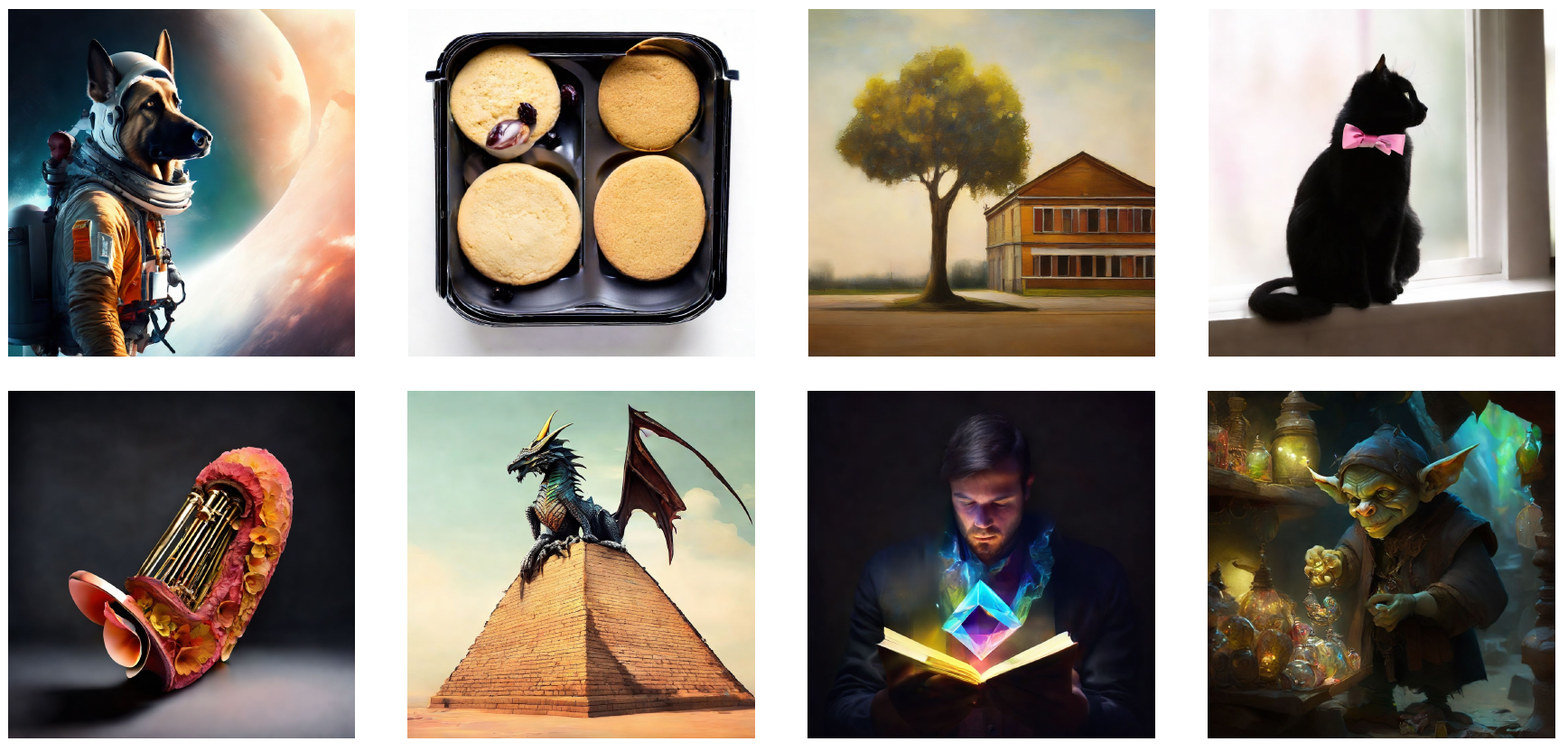}
    \caption{Detailed qualitative results: 1024px samples from \model, Part 1. 
    }
    \label{fig:t2i_1024px_results_1}
\end{figure}

\begin{figure}[htbp]
    \vspace{0em}
    \centering
    \includegraphics[width=0.99\textwidth]{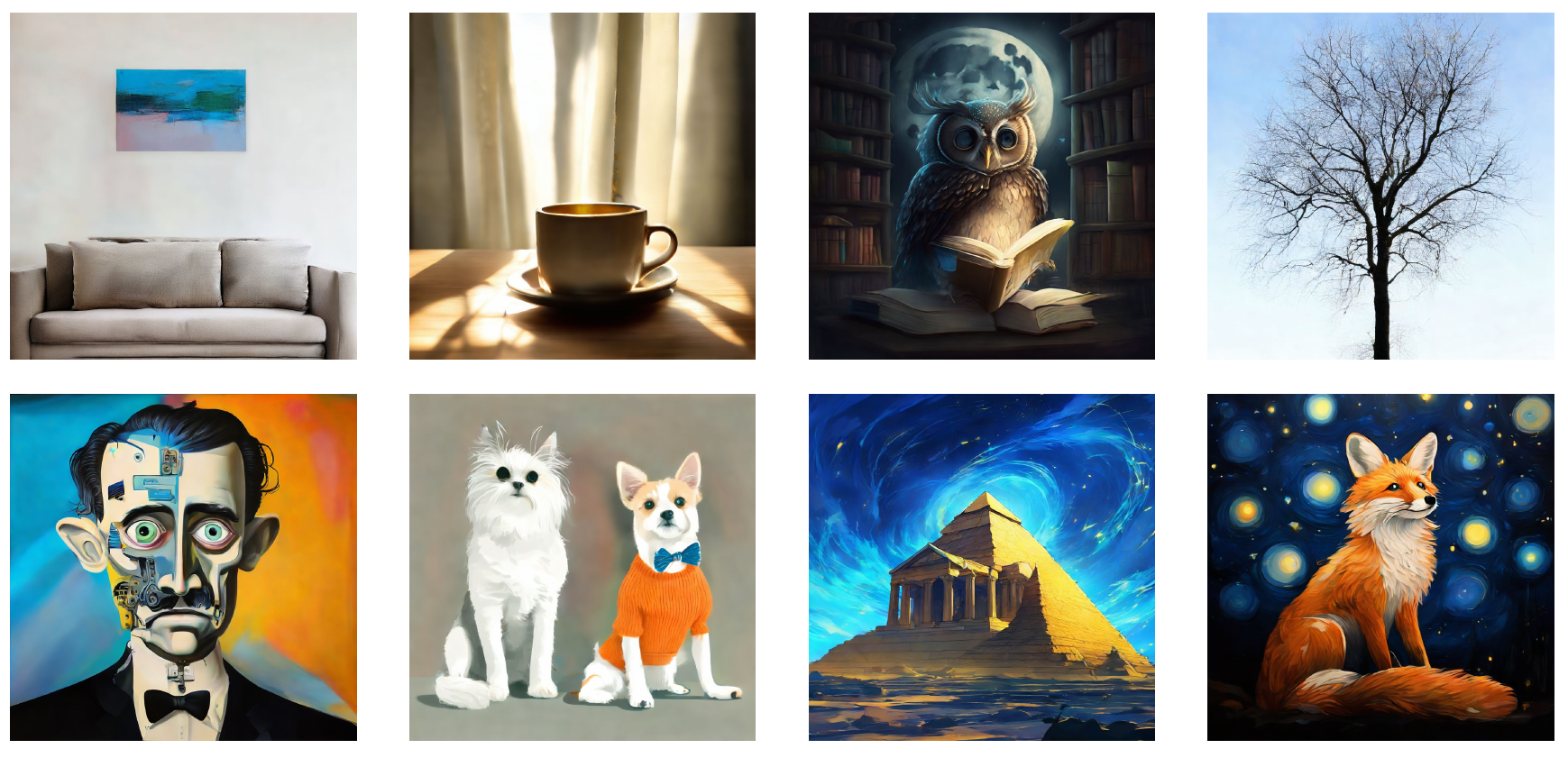}
    \caption{Detailed qualitative results: 1024px samples from \model, Part 2. 
    }
    \label{fig:t2i_1024px_results_2}
\end{figure}

As illustrated in Fig.~\ref{fig:t2i_1024px_results_1} and \ref{fig:t2i_1024px_results_2}, we provide additional qualitative results sampled from 1024px images generated by \model. 
\model produces high-fidelity images with convincing details and textures. These results support our belief that the proposed \model offers both practical effectiveness and potential value for generative modeling. 

\end{document}